\pdfoutput=1
\documentclass[10pt,twocolumn,letterpaper]{article}

\usepackage[pagenumbers]{cvpr}

\usepackage{makecell}
\usepackage[T1]{fontenc}
\usepackage{nicefrac}
\usepackage{microtype}
\usepackage{multirow}
\usepackage{float}
\usepackage{dsfont}
\newcommand{\mathbbm}[1]{\mathds{#1}}
\newcounter{algorithm}

\definecolor{cvprblue}{rgb}{0.21,0.49,0.74}
\usepackage[pagebackref,breaklinks,colorlinks,allcolors=cvprblue]{hyperref}

\newcommand{\bigfig}[2][]{%
  \makebox[\linewidth][c]{\includegraphics[width=\textwidth,height=.86\textheight,keepaspectratio,#1]{#2}}}

\AddToShipoutPicture{%
  \AtTextUpperLeft{%
    \put(0,\LenToUnit{20pt}){\makebox[\textwidth][l]{\small Jang \textit{et al.}}}%
    \put(0,\LenToUnit{20pt}){\makebox[\textwidth][r]{\small MAST}}%
  }%
}

\title{MAST: Spiking Neural Networks for AI-Generated Video Detection\\via Temporal Residuals}

\author{Minsuk Jang$^{1}$\hspace{0.6em} Yujin Yang$^{1}$\hspace{0.6em} Hee-Seon Kim$^{1}$\hspace{0.6em} Minseok Son$^{1}$\hspace{0.6em} Younghun Kim$^{1}$\hspace{0.6em} Changick Kim$^{1,\dagger}$\\[4pt]
$^{1}$Korea Advanced Institute of Science and Technology (KAIST), Daejeon, Korea\\[2pt]
{\normalsize $^{\dagger}$Corresponding author}}

\begin{document}
\maketitle

\begin{abstract}
Modern AI-generated videos are photorealistic at the single-frame level, making temporal dynamics an important axis for detection. Existing detectors exploit temporal cues, but the type and timescale of these cues vary across generators. A caption-paired analysis reveals smoother pixel-level residual dynamics and more compact semantic trajectories in generated clips than in real clips. Motivated by the event-driven dynamics of spiking neural networks (SNNs), we recast temporal residual sequences from RGB videos as multi-channel pseudo-events. We propose MAST, which learns per-channel integration timescales for these pseudo-events and fuses the resulting temporal features with semantic features and trajectory statistics from a frozen video encoder. On GenVideo, MAST achieves 93.87\% mean AUROC across ten generators unseen during training. In ablations, MAST achieves higher mean AUROC than the raw-RGB and single-branch variants. With pseudo-event inputs, its spiking temporal module yields detection performance comparable to that of an ANN temporal module of equal depth and width.
\end{abstract}

\begin{figure*}[t]
  \centering
  \includegraphics[width=.85\linewidth]{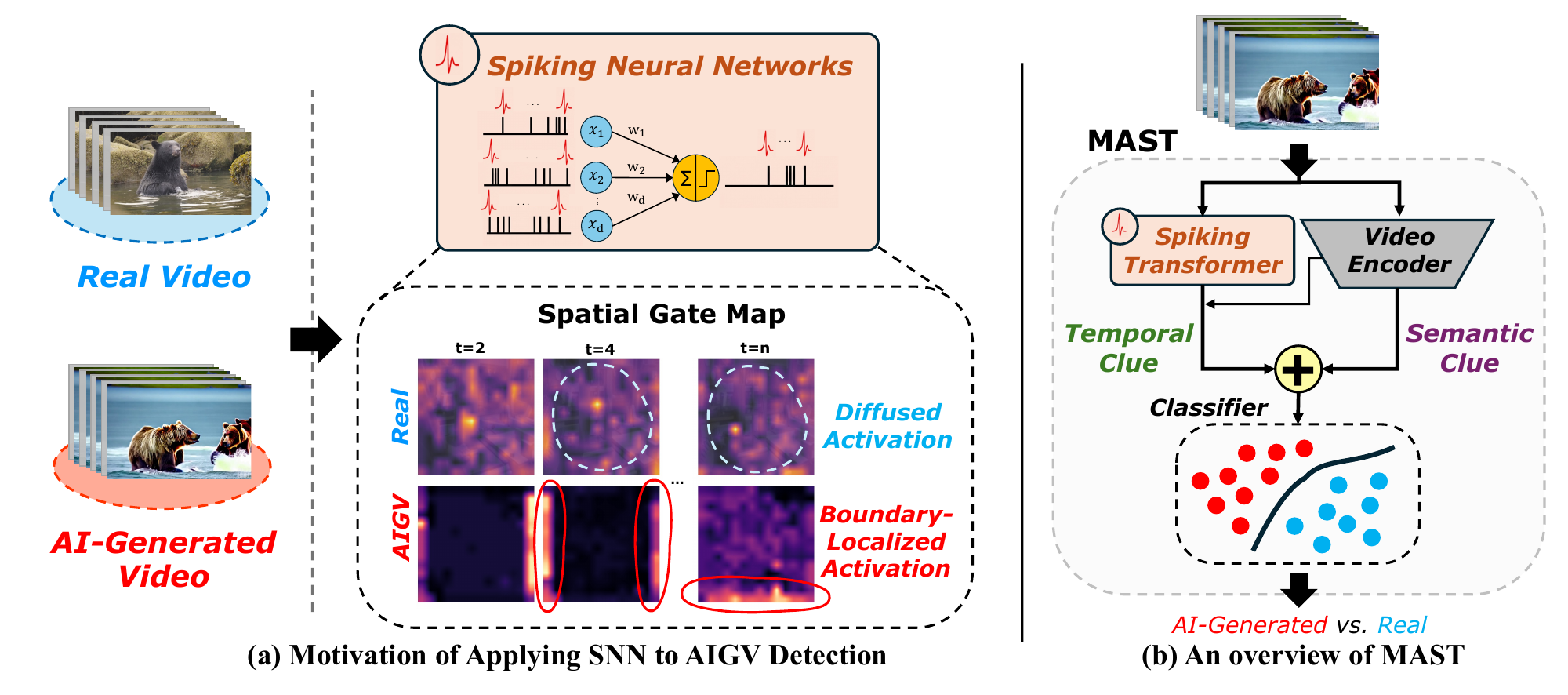}
  \caption{(a) Motivation. For the clips shown, the SNN response is more diffuse in the real example and more concentrated near the image border in the generated example. (b) MAST
  overview. Motivated by this observation, MAST combines spike-based
  temporal integration over pseudo-events with a semantic
  trajectory encoder, uses trajectory curvature as an auxiliary cue,
  and fuses temporal and semantic features.}
  \label{fig:teaser}
\end{figure*}

\section{Introduction}
\label{sec:intro}

AI-generated videos (AIGVs) have rapidly approached frame-level
photorealism~\cite{sora,blattmann2023stable,hong2022cogvideo,
chen2024videocrafter2}. A video, however, is more than a set of
images: it unfolds along a time axis, which offers a detection cue
that a single image lacks. As frame-level artifacts fade, this
temporal cue becomes increasingly important. Early work identified
traces left by video generators within individual
frames~\cite{Vahdati_2024_CVPR}, and subsequent studies showed that
changes in frame representations over time provide an additional
detection cue~\cite{zheng2025d3,interno2026restrav}.

Prior detectors already exploit temporal cues through frame
differences, motion residuals, feature trajectories, and higher-order
statistics~\cite{ma2025detecting,xu2023tall,interno2026restrav,
chen2024demamba,zheng2025d3,zhang2025nsgvd}. These approaches use
temporal backbones~\cite{ma2025detecting,chen2024demamba}, fixed
temporal descriptors~\cite{zheng2025d3,interno2026restrav}, or
statistical discrepancy measures~\cite{zhang2025nsgvd}. D3~\cite{zheng2025d3} and
ReStraV~\cite{interno2026restrav}, for example, do not learn a separate temporal integration
timescale for each residual cue. The magnitude and timescale of a given temporal cue can vary across
generators, as can the cues that are most informative for detection. This diversity can make
detectors built around a single cue or a fixed temporal summary less
robust under cross-generator evaluation. 

Spiking neural networks (SNNs) have been applied to video understanding~\cite{zou2025spikevideoformer}, but their use for detecting fully AI-generated videos has received limited attention.

Two observations motivate our design. \textbf{(1) Spatial gate responses.} In Figure~\ref{fig:teaser}(a), the spatial gate maps show diffuse activity in the real example and activity concentrated near the image border in the generated example. \textbf{(2) Temporal signatures.} In GenVidBench~\cite{ni2026genvidbench} Pair 1, four video generators share the same set of text prompts. Using real videos from Vript as a common reference, we ask which temporal differences recur across these generators. All four exhibit lower residual spectral centroids and more compact trajectories in X-CLIP feature space~\cite{xclip} relative to the Vript reference. These findings motivate a detector that preserves residual evolution over time and combines pixel-level dynamics with semantic trajectory cues.

To integrate residual cues over time, we turn to SNNs, whose membrane potentials retain information from earlier inputs~\cite{fang2021plif}. Small changes in intensity that a continuous ANN activation could represent may yield the same spike output at a given frame. We therefore make local changes explicit at the input. For example, a bright stationary region has high RGB intensity across frames but a near-zero temporal residual. Drawing on event cameras, which respond to local brightness changes~\cite{gallego2020event,hu2021v2e}, we compute temporal residual maps from RGB frames and apply a soft threshold to generate \emph{pseudo-events} (Figure~\ref{fig:comparison}). These maps encode where and when changes occur, allowing the SNN to integrate successive responses over a learned timescale for each channel.

Building on this match, we introduce \textbf{MAST}
(\textbf{M}ulti-channel pseudo-event SNN with \textbf{A}daptive
\textbf{S}piking \textbf{T}emporal integrators;
Figure~\ref{fig:teaser}), which combines temporal and semantic
components. The temporal component represents multiple residual cues
as pseudo-events and integrates each channel with a learnable
timescale. The semantic component uses a frozen X-CLIP~\cite{xclip}
encoder to provide semantic features and trajectory statistics,
which are fused with the temporal features from the spiking branch.

Our contributions are:
\begin{itemize}
    \interlinepenalty=10000%
    \item \textbf{Analysis.} On GenVidBench Pair~1, all four generators exhibit lower residual spectral centroids and smaller semantic trajectory volumes than the real videos they are compared against.

    \item \textbf{Method.} To our knowledge, \textbf{MAST} is the first
    SNN-based detector for fully AI-generated videos. MAST represents
    multiple temporal residual cues as pseudo-events, integrates them
    with per-channel learnable timescales, and fuses the resulting
    temporal features with frozen semantic features and trajectory
    statistics.

    \item \textbf{Empirical evaluation.} On GenVideo~\cite{chen2024demamba}, MAST achieves 93.87\% mean AUROC and 92.41\% mean fake recall across ten unseen generators, competitive with recent ANN-based detectors.
\end{itemize}

\begin{figure*}[t]
  \centering
  \includegraphics[width=\linewidth]{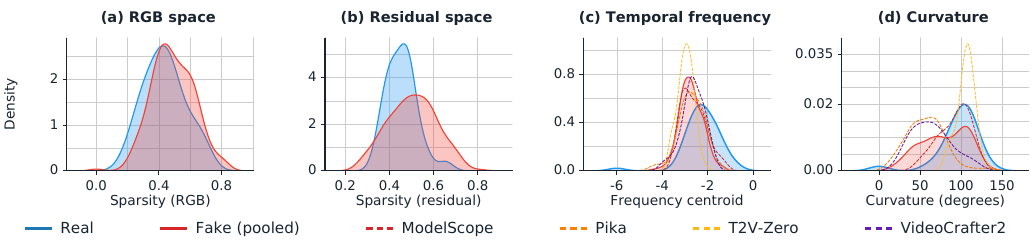}
  \caption{GenVidBench Pair 1. Distributions of (a) RGB sparsity,
  (b) Laplacian residual sparsity, (c) spectral centroid $f_c$, and
  (d) angular curvature $\theta$. Real and fake distributions overlap,
  with shifts in residual and temporal statistics that vary by generator.}
  \label{fig:dist}

\end{figure*}

\section{Related Work}

\subsection{Temporal Cues for Detection}
\label{sec:rw_video}
Existing detectors use temporal cues such as motion estimates and
changes in frame features to identify AI-generated videos. AIGVDet~\cite{bai2024aigvdet} uses optical-flow
anomalies, while MM-Det~\cite{song2024mmdet} combines multimodal forgery
representations with spatiotemporal attention and feature fusion.
GenVideo~\cite{chen2024demamba}, GenVidBench~\cite{ni2026genvidbench},
and IvyFake~\cite{jiang2025ivyfake} evaluate detection across generators
and data sources.

Within this literature, methods differ in how they aggregate temporal
evidence. DeCoF~\cite{ma2025detecting} applies a learned Transformer to
frame features, whereas DeMamba~\cite{chen2024demamba} models local
spatiotemporal feature consistency with a Mamba module.
D3~\cite{zheng2025d3} measures variability in second-order feature
differences. ReStraV~\cite{interno2026restrav} combines stepwise distances
and curvatures with their summary statistics for classification.
NSG-VD~\cite{zhang2025nsgvd} measures distributional discrepancies
between normalized spatiotemporal-gradient features of test and real
videos using maximum mean discrepancy. MAST shares their focus on
temporal evidence, but \textbf{constructs multi-channel pseudo-events
from temporal residuals and integrates each cue with a learnable
spiking timescale} before fusing the resulting features with frozen
semantic features and trajectory statistics.

\subsection{Spiking Neural Networks}
\label{sec:rw_snn}
Spiking Neural Networks (SNNs)~\cite{maass1997snn} encode information
through spikes while maintaining neuronal states that evolve over
time. The Leaky Integrate-and-Fire (LIF) neuron integrates inputs
through a membrane state controlled by a time constant $\tau$.
PLIF~\cite{fang2021plif} learns this time constant, while
DIET-SNN~\cite{rathi2021diet} jointly optimizes membrane leak and
firing thresholds. These mechanisms provide a basis for adapting
integration timescales to different residual cues.

Event cameras encode local brightness changes as asynchronous
events~\cite{gallego2020event}, providing a natural input representation
for spike-based processing. Software such as v2e~\cite{hu2021v2e}
synthesizes event streams from conventional video frames.
Spike-driven Transformers~\cite{yao2023spike,yao2025sdtv3} provide
attention-based SNN architectures, and
SpikeVideoFormer~\cite{zou2025spikevideoformer} models spatiotemporal
features for video understanding. STSep~\cite{li2025stsep} separates
spatial and temporal processing and uses explicit temporal differences
to capture motion. UCF-Crime-DVS~\cite{qian2025ucf} studies SNN-based
anomaly detection in event videos. These studies establish temporal
processing in SNNs; MAST targets the authenticity of AI-generated
videos through forensic residual cues and semantic representations.

\begin{figure}[h!]
  \centering
  \includegraphics[width=0.9\linewidth]{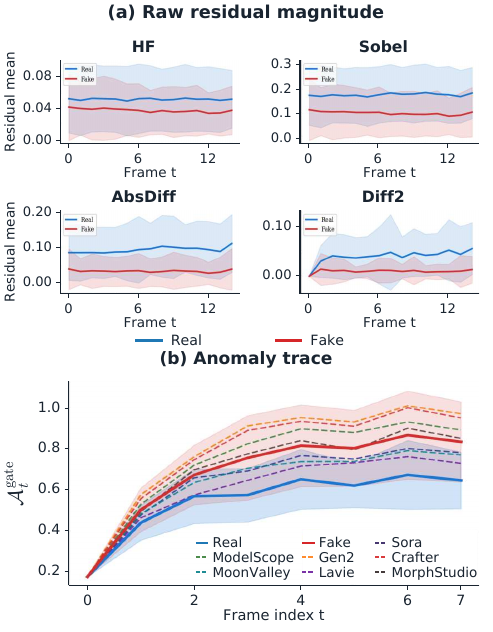}
\caption{(a) Per-frame spatial mean $\pm\sigma$ of the temporal
residual, real (blue) vs.\ fake (red, all generators pooled);
fake means are lower over the displayed residual sequence.
(b) Eight-step anomaly trace $\mathcal{A}^{\mathrm{gate}}_t$ per
generator: the LIF potential driven by the gate response of the
trained temporal branch.}
  \label{fig:temporal_anomaly}
\end{figure}

\section{Temporal Signatures in AI-generated Videos}
\label{sec:motivation}

We compare four generators in GenVidBench Pair~1 with a common reference of real videos from Vript. Two temporal signatures recur across all four generators: \textbf{lower residual spectral centroids} and \textbf{smaller semantic trajectory volumes} in generated clips relative to the real-video reference.

\noindent\textbf{GenVidBench.} In Pair~1 of GenVidBench~\cite{ni2026genvidbench}, ModelScope~\cite{wang2023modelscope}, Pika~\cite{pika2024}, Text2Video-Zero~\cite{khachatryan2023text2video}, and VideoCrafter2~\cite{chen2024videocrafter2} share the same set of text prompts. We compare each generator with the same set of real videos from Vript~\cite{yang2024vript}.

\begin{figure*}[ht!]
  \centering
  \includegraphics[width=0.77\linewidth,trim={0 40 0 0},clip]{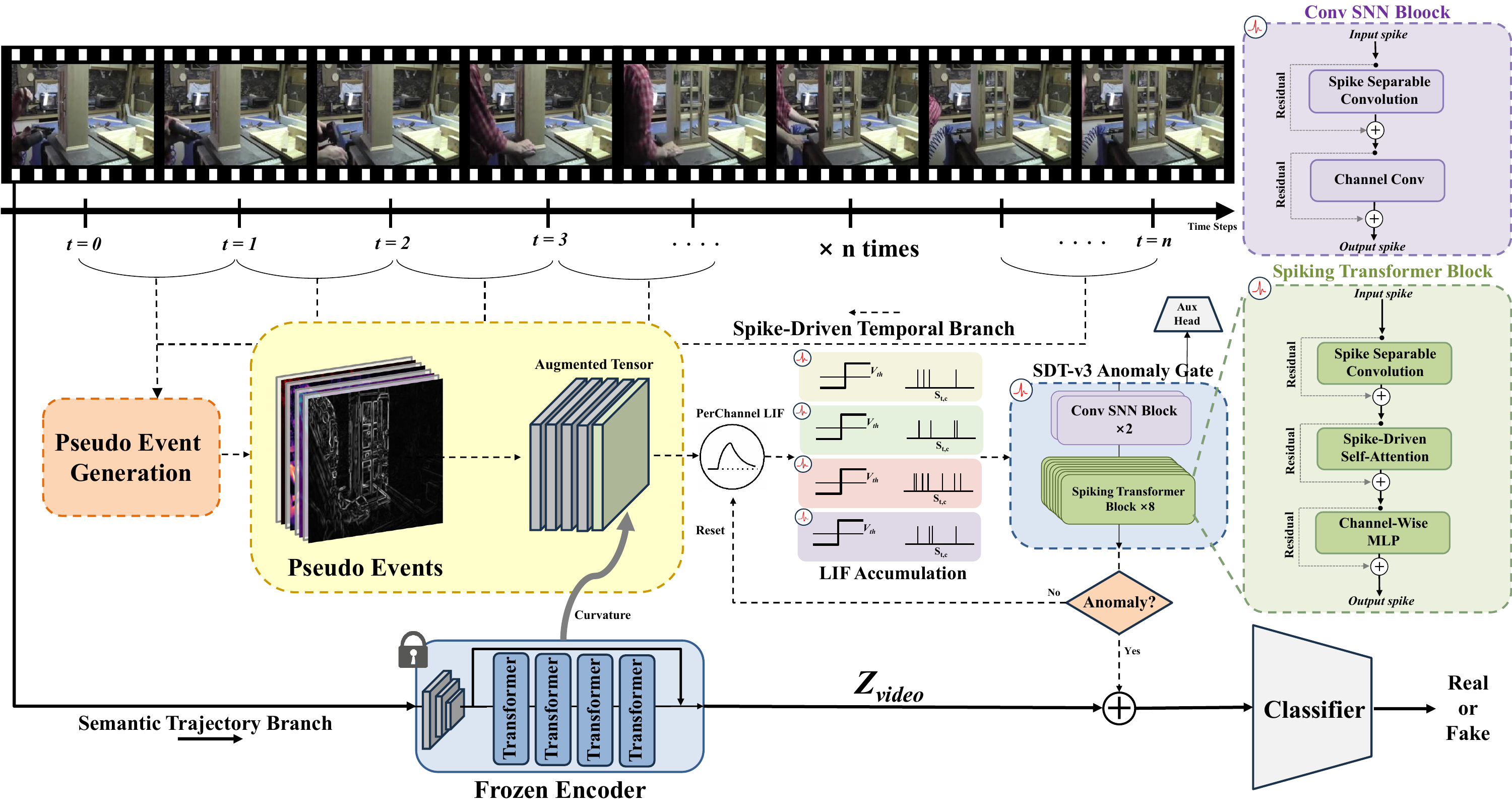}
    \caption{MAST overview. A trainable Spike-Driven Temporal
    Branch (SDTB, dashed) processes multi-channel pseudo-event
    residuals; a frozen Semantic Trajectory Branch (solid) provides
    a video-level semantic feature and trajectory geometry. The
    classifier fuses $Z_{\mathrm{video}}$, $F_{\mathrm{anom}}$,
    $F_{\mathrm{gate}}$, and trajectory statistics $F_{\mathrm{traj}}$.}
  \label{fig:method_overview}
\end{figure*}

\subsection{Pixel and Semantic Signatures}
\label{sec:structure}

\begin{table}[t]
  \centering
  
  \caption{Statistics on GenVidBench (Pair 1).
  $\mathcal{S}_\mathrm{rgb}$, $\mathcal{S}_\mathrm{res}$: Hoyer sparsity
  in RGB and high-frequency residual space; $f_c$: spectral centroid
  of the residual signal; Vol.: convex-hull volume of the
  trajectory.}
  \label{tab:per_gen_stats}
  
  {\footnotesize
  \setlength{\tabcolsep}{4pt}
  \begin{tabular}{lcccc}
    \toprule
    \textbf{Generator} & $\mathcal{S}_\mathrm{rgb}$ & $\mathcal{S}_\mathrm{res}$ & $f_c$ & Vol. \\
    \midrule
    Real (Vript)        & $0.438$ & $0.506$ & $0.0178$ & $\mathbf{19.07}$ \\
    \midrule
    ModelScope          & $0.483$ & $0.559$ & $0.0057$ & $5.54$ \\
    Pika                & $0.466$ & $0.447$ & $0.0040$ & $9.94$ \\
    Text2Video-Zero     & $0.455$ & $0.582$ & $0.0014$ & $8.13$ \\
    VideoCrafter2       & $0.504$ & $0.312$ & $0.0038$ & $3.23$ \\
    \bottomrule
  \end{tabular}
  }
  
\end{table}

Figure~\ref{fig:dist} and Table~\ref{tab:per_gen_stats} summarize pixel-level residual statistics and semantic trajectory geometry. Hoyer sparsity~\cite{hoyer2004non} measures the spatial concentration of RGB intensities ($\mathcal{S}_\mathrm{rgb}$) and inter-frame changes in Laplacian responses~\cite{burt1983laplacian} ($\mathcal{S}_\mathrm{res}$). The residual spectral centroid $f_c$ is the energy-weighted mean temporal frequency. Higher values indicate a greater contribution from faster changes. In X-CLIP feature space, angular curvature measures the change in direction between consecutive displacement vectors. Convex-hull volume measures the space occupied by a clip's trajectory after projection onto three shared principal components.

Let $Y_t$ denote the luma of frame $t$ and $\mathrm{Lap}$ the spatial Laplacian. RGB and residual sparsity statistics ($\mathcal{S}_\mathrm{rgb}$ and
$\mathcal{S}_\mathrm{res}$, with $\Delta^{\mathrm{HF}}_t = |\mathrm{Lap}(Y_t)
- \mathrm{Lap}(Y_{t-1})|$) do not separate real from fake
(Figure~\ref{fig:dist}(a, b)). The temporal cues, however, show clearer separation: the
spectral centroid $f_c$ and the angular curvature/volume of the
X-CLIP trajectory both place generators on the same side of the real-video
reference (Figure~\ref{fig:dist}(c, d)), with $f_c$ $3\times$ to
$13\times$ lower for generators than for Real and convex-hull volume
$2\times$ to $6\times$ smaller. Together, these pixel-level and semantic-space gaps point to a
common latent-interpolation-driven temporal smoothness in
generated video.

\subsection{Residual Dynamics}
\label{sec:dynamics}

We next examine how the residual signal evolves over time.
Figure~\ref{fig:temporal_anomaly}(a) shows lower fake means than real
means in the four displayed channels over the plotted residual sequence.
These changes in residual magnitude motivate processing the residual maps in temporal order before computing a clip-level summary.

To summarize these dynamics in a single time-varying scalar, we use
the trained spiking temporal branch of MAST
(Section~\ref{sec:spiking_branch}), whose gate head outputs a spatial
anomaly map $G_t$ at each timestep. We take its spatial mean $g_t$
and accumulate it through a non-firing leaky integrator with a fixed
leak constant $\tau_{\mathrm{anom}}$:
\begin{equation}
    \mathcal{A}^{\mathrm{gate}}_t = \mathcal{A}^{\mathrm{gate}}_{t-1} \exp\!\left(-\frac{1}{\tau_{\mathrm{anom}}}\right) + g_t.
    \label{eq:anomaly}
\end{equation}
On GenVideo test sources, every fake source accumulates a higher
potential than real clips after the first timestep
(Figure~\ref{fig:temporal_anomaly}(b)), even though raw residual
magnitudes in (a) are lower for fake clips: the trace reflects learned
anomaly evidence, not residual energy.

\section{MAST Framework}
\label{sec:method}

The observations in Section~\ref{sec:motivation} lead directly
to three design choices for MAST. \textbf{First}, the
$f_c$ and trajectory-volume gaps in
Section~\ref{sec:structure} show that discriminative evidence
unfolds over time, so temporal residuals must not be collapsed
to a fixed scalar or descriptor before classification. \textbf{Second}, the residual channels in
Section~\ref{sec:dynamics} (Figure~\ref{fig:temporal_anomaly}(a))
differ in magnitude and temporal profile, so temporal integration
must be adaptive across cues rather than governed by a single shared
time constant. \textbf{Third}, the
$\mathcal{S}_\mathrm{rgb}$/$\mathcal{S}_\mathrm{res}$ failures
contrasted with the cleaner $f_c$ and trajectory-volume
separation show that pixel-level and semantic-level evidence are
complementary, since generated videos can resemble real videos in one space
while remaining inconsistent in the other, so a single
modality is insufficient. MAST instantiates these three
requirements through two complementary pathways
(Figure~\ref{fig:method_overview}, Algorithm~\ref{alg:mast}): a trainable
\textbf{\emph{Spike-Driven Temporal Branch} (SDTB)} that
processes pixel-level residuals as pseudo-events, and a frozen
\textbf{\emph{Semantic Trajectory Branch}} that produces both
the geometry of each video's trajectory in the latent feature
space and a representative video-level semantic feature.

\begin{figure}[h]
  \centering
  
  \includegraphics[width=\linewidth]{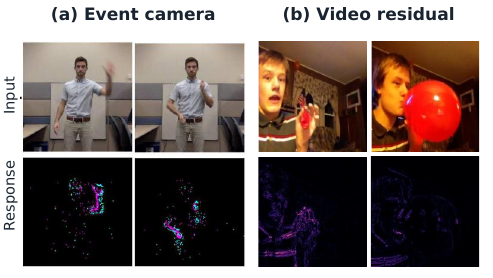}
  \caption{(a) Event-camera frames from DVS-Gesture~\cite{dvsgesture}.
  (b) Pseudo-events derived from a conventional video. In both cases, responses are concentrated at pixels that change over time, supporting temporal residuals as event-like inputs to the temporal module.
  }
  \label{fig:comparison}

\end{figure}

\subsection{Residual Pseudo-Events}
\label{sec:motiv_residual}

We represent inter-frame residuals as event-like pseudo-events (Figure~\ref{fig:comparison}) so
that the SDTB operates on changes rather than raw
appearance. Let $F_t \in \mathbb{R}^{H \times W \times 3}$ denote the
video frame at time $t$. We use \emph{temporal residual} for a quantity computed from changes across frames in pixel or feature space. The simplest example is the elementwise absolute RGB difference,
\begin{equation}
\Delta F_t = \big|F_t-F_{t-1}\big|.
\label{eq:residual}
\end{equation}
Residuals are computed on raw RGB frames and rescaled by a per-frame
mean before thresholding (Appendix~\ref{app:postproc}). Averaging $\Delta F_t$ over RGB channels gives the AbsDiff map. HF, Sobel, and Chroma measure inter-frame changes in Laplacian responses, edge magnitudes, and chroma components, respectively. Diff2 measures the absolute difference between consecutive AbsDiff maps. Temporal differences can also be computed in feature space~\cite{zheng2025d3,interno2026restrav}. These signals are sparse and
concentrated at pixels that change between frames.

\begin{figure*}[t]
\centering
\includegraphics[width=\textwidth]{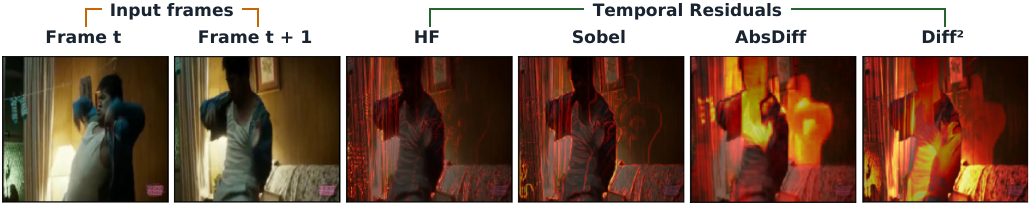}
\caption{Input frames $F_t$ and $F_{t+1}$ and four pixel-level
temporal residual channels: high-frequency Laplacian (HF), Sobel
edge magnitude difference, absolute frame difference (AbsDiff),
and second-order frame difference (Diff$^2$). Diff$^2$ uses three
consecutive frames; only two input frames are shown.}
\label{fig:temporal_residuals}
\end{figure*}

This structure is closely aligned with the event-driven regime in
which SNNs operate naturally~\cite{gallego2020event,hu2021v2e,
maass1997snn,rossello2022snn}. Let $I_t(x,y) \in \mathbb{R}_{>0}$
denote the pixel intensity at spatial location $(x,y)$ at time
$t$. An event camera triggers a binary event at $(x,y)$ whenever
the log-intensity change between consecutive frames exceeds a
contrast threshold $c \in \mathbb{R}_{>0}$,
\begin{equation}
e_t(x,y) \;=\; \mathbbm{1}\!\left[\,\big|\log I_t(x,y) - \log I_{t-1}(x,y)\big| \geq c\,\right],
\label{eq:event_cam}
\end{equation}
and we adopt the same structural form, but replace the hard
threshold with a differentiable soft one:
\begin{equation}
E_t(x,y) \;=\; \sigma\!\left(\frac{\Delta F_t(x,y) - c_{\mathrm{th}}}{\beta}\right),
\label{eq:pseudo_event}
\end{equation}
where $\sigma(\cdot)$ is the logistic function,
$c_{\mathrm{th}} \in \mathbb{R}_{>0}$ is a fixed threshold,
and $\beta > 0$ is a temperature controlling the sharpness of
the soft threshold. All main results use seven channels: five pixel
residuals (HF, Sobel, Chroma, AbsDiff, Diff2; Figure~\ref{fig:temporal_residuals})
and two semantic trajectory channels (displacement, curvature).

\subsection{Adaptive Spiking Integration}
\label{sec:spiking_branch}
\label{sec:prelim_lif}
\label{sec:motiv_snn}

This module constructs a multi-channel pseudo-event stream $E \in
\mathbb{R}^{T \times C \times H' \times W'}$ from five pixel-level
residual channels (HF, Sobel, Chroma, AbsDiff, Diff2)
and two semantic-level residual channels derived from the frozen patch tokens
$Z_{\mathrm{patch}} \in \mathbb{R}^{T \times N \times D}$ of the
semantic trajectory encoder (Section~\ref{sec:semantic_branch}),
where $T$ is the number of frames in the clip, $N$ is the number of
patch tokens per frame, and $D$ is the embedding dimension. We
write $Z^{t,n}_{\mathrm{patch}} \in \mathbb{R}^{D}$ for the token at
frame $t \in \{1, \ldots, T\}$ and patch index $n \in \{1, \ldots,
N\}$, and define the per-patch temporal difference $\Delta
Z^{t,n}_{\mathrm{patch}} = Z^{t,n}_{\mathrm{patch}} -
Z^{t-1,n}_{\mathrm{patch}}$. The two semantic channels are the
per-patch displacement $d^{t}_{n}$ and trajectory curvature
$\kappa^{t}_{n}$,
\begin{align}
d^{t}_{n}     &= \big\| \Delta Z^{t,n}_{\mathrm{patch}} \big\|_2,
  \label{eq:traj_disp}\\
\kappa^{t}_{n} &= \arccos\!\frac{
  \big\langle \Delta Z^{t,n}_{\mathrm{patch}},\,
              \Delta Z^{t-1,n}_{\mathrm{patch}} \big\rangle}{
  \big\| \Delta Z^{t,n}_{\mathrm{patch}} \big\|_2 \,
  \big\| \Delta Z^{t-1,n}_{\mathrm{patch}} \big\|_2}.
  \label{eq:traj_curv}
\end{align}

The seven channels carry information at different temporal scales.
Following prior work on learnable neuronal dynamics~\cite{fang2021plif,rathi2021diet},
we use channel-wise LIF
neurons with independently learnable time constants $\tau_c$ and
firing thresholds $V_{\mathrm{th},c}$, both parameterized in the
log-domain ($\tau_c = 2 \cdot \exp(\theta_{\tau, c})$, $V_{\mathrm{th},c} = \exp(\theta_{V, c})$), which we
refer to as \textbf{PerChannelLIF}. Its integration, multispike firing,
and soft reset are
\begin{equation}
\begin{aligned}
v_t^{(c)} &= \big(1-\tfrac{1}{\tau_c}\big)v_{t-1}^{(c)}+x_t^{(c)},\\
s_t^{(c)} &= \Big\lfloor\mathrm{clamp}\!\big(v_t^{(c)}/V_{\mathrm{th},c},0,4\big)+\tfrac{1}{2}\Big\rfloor,\\
v_t^{(c)} &\leftarrow v_t^{(c)}-s_t^{(c)}V_{\mathrm{th},c}.
\end{aligned}
\label{eq:perchannel_main}
\end{equation}
Here, $x_t^{(c)}$ is the input current, $v_t^{(c)}$ the membrane
potential, and $s_t^{(c)}\in\{0,1,2,3,4\}$ the spike output.
Learning $\tau_c$ adapts memory retention and the integration timescale
of each residual channel. Training details and the
full forward algorithm appear in Appendix~\ref{app:training_details}.

Channel-wise spikes from PerChannelLIF are processed by an
attention block built on the Efficient Spike-Driven
Self-Attention (E-SDSA) topology of
SDT-V3~\cite{yao2025sdtv3, yao2023spike}, in which queries,
keys, and values $Q_S, K_S, V_S$ are spike tensors over the
(token, time) joint dimension. The attention output is computed
as
\begin{equation}
\begin{split}
\mathrm{SpikeAttn}(Q_S,K_S,V_S) \;=\;& \\
\mathrm{Linear}_{out}\Big(\mathrm{Multispike}\big(& (Q_S K_S^\top V_S) \cdot \mathrm{scale}\big)\Big),
\end{split}
\label{eq:spikeattn}
\end{equation}
where $\mathrm{scale}$ is the standard attention scaling factor,
$\mathrm{Multispike}(\cdot)$ is the $L$-level multispike firing
function (Appendix~\ref{app:training_details},
Eq.~\ref{eq:multispike_forward}), and $\mathrm{Linear}_{out}$ is
the output projection. The gate head produces a spatial map $G_t\in\mathbb{R}^{H'\times W'}$ at each timestep. We summarize each map by its spatial mean and maximum. The leaky integrator of Eq.~\ref{eq:anomaly} accumulates the spatial
mean over time to produce the anomaly trace
$\mathcal{A}^{\mathrm{gate}}_t$.

\subsection{Semantic Trajectory Encoding}
\label{sec:semantic_branch}
\label{sec:motiv_trajectory}

A frozen X-CLIP encoder~\cite{xclip} produces frame embeddings $Z_t\in\mathbb{R}^D$ in the temporal context of the clip through cross-frame attention. These embeddings define a semantic trajectory and its per-step displacement,
\begin{equation}
\mathcal{T}=(Z_1,Z_2,\ldots,Z_T), \qquad \Delta Z_t=Z_t-Z_{t-1}.
\label{eq:trajectory}
\end{equation}
The clip embedding $Z_{\mathrm{video}}$ enters the fused representation directly. The frame embeddings provide the trajectory statistics $F_{\mathrm{traj}}$. From the patch tokens $Z_{\mathrm{patch}}\in\mathbb{R}^{T\times N\times D}$, we compute displacement and curvature at each patch location. These two channels are inputs to PerChannelLIF.

\subsection{Fusion and Training}
\label{sec:fusion}
We project the SDTB anomaly trace $\mathcal{A}^{\mathrm{gate}}$ and maps $G_t$ into $F_{\mathrm{anom}}$ and $F_{\mathrm{gate}}$. To form $F_{\mathrm{traj}}$, we measure the distance between successive frame embeddings and the absolute difference between successive distances. For both Euclidean and cosine distance, we summarize each of the two sequences by its mean, standard deviation, maximum, and minimum (Appendix~\ref{app:trajectory_summary}). We concatenate
\[
F_{\mathrm{final}} = [Z_{\mathrm{video}} \parallel F_{\mathrm{anom}} \parallel F_{\mathrm{gate}} \parallel F_{\mathrm{traj}}].
\]
A lightweight MLP predicts $\hat{y}=\phi(F_{\mathrm{final}})$.
We use three task losses and a spike-rate regularizer:
\begin{equation}
\begin{split}
    \mathcal{L} ={}& \mathcal{L}_{\mathrm{BCE}}(\hat{y}, y)
    + \lambda_1 \mathcal{L}_{\mathrm{BCE}}(\hat{y}_{\mathrm{snn}}, y)\\
    &+ \lambda_2 \mathcal{L}_{\mathrm{supcon}} + \mathcal{L}_{\mathrm{rate}}.
\end{split}
\end{equation}
The rate term is defined in Eq.~\ref{eq:rate_penalty}.
The first term, $\mathcal{L}_{\mathrm{BCE}}(\hat{y}, y)$, is the
main binary cross-entropy on the fused prediction $\hat{y}$ from
$F_{\mathrm{final}}$. The second term,
$\mathcal{L}_{\mathrm{BCE}}(\hat{y}_{\mathrm{snn}}, y)$, is an
auxiliary binary cross-entropy on $\hat{y}_{\mathrm{snn}}$, the
prediction of an auxiliary classification head that operates on the
SDTB features alone; this term forces the SDTB to be a
self-sufficient temporal-cue classifier rather than a residual
offset on top of the dominant frozen X-CLIP feature. The third term,
$\mathcal{L}_{\mathrm{supcon}}$, applies supervised contrastive
learning~\cite{khosla2020supcon} on $F_{\mathrm{final}}$ to pull
same-label clips together and push real and generated clips apart
in the fused representation space. To handle the
non-differentiable spike function during backpropagation, we use
the ATan surrogate
gradient~\cite{neftci2019surrogate, fang2023spikingjelly}; full
training details are deferred to
Appendix~\ref{app:training_details}.

\begin{table*}[t!]
\centering
\caption{GenVideo cross-generator results (\%) under the Kinetics-400/Pika protocol.
R: fake recall; BAcc: balanced accuracy, i.e., the mean of fake recall and
real-video specificity; AUROC: area under the ROC curve;
Avg.: generator-wise mean. Published baseline results follow~\cite{zhang2025nsgvd}, Table~1,
and use class-balanced test subsets;
$\dagger$: our reevaluations on the full test set, which share one real reference set
across generators;
Published and reevaluated results retain their respective test samples and operating points.}
\label{tab:genvideo_pika}
{\fontsize{7.5}{8}\selectfont
\setlength{\tabcolsep}{2.5pt}
\renewcommand{\arraystretch}{0.95}
\setlength{\aboverulesep}{0.25ex}
\setlength{\belowrulesep}{0.35ex}
\begin{tabular*}{\textwidth}{@{\extracolsep{\fill}}c|c|cccccccccc|c@{}}
\toprule
\textbf{Method} & \textbf{Metric} & \textbf{ModelScope} & \textbf{MorphStudio} & \textbf{MoonValley} & \textbf{HotShot} & \textbf{Show1} & \textbf{Gen2} & \textbf{Crafter} & \textbf{Lavie} & \textbf{Sora} & \textbf{WildScrape} & \textbf{Avg.} \\
\midrule
\multirow{3}{*}{DeMamba~\cite{chen2024demamba}}
& R & 87.00 & 93.60 & 98.80 & 40.60 & 48.40 & 98.00 & 88.40 & 59.00 & 48.21 & 58.20 & 72.02 \\
& BAcc & 91.70 & 95.00 & 97.60 & 68.50 & 72.40 & 97.20 & 92.40 & 77.70 & 72.32 & 77.30 & 84.21 \\
& AUROC & 98.04 & 98.82 & 99.68 & 87.84 & 90.12 & 99.46 & 97.81 & 91.32 & 88.36 & 87.38 & 93.88 \\
\midrule
\multirow{3}{*}{NPR~\cite{tan2024npr}}
& R & 61.20 & 80.00 & 98.00 & 16.00 & 33.00 & 91.20 & 80.60 & 34.60 & 35.71 & 43.20 & 57.35 \\
& BAcc & 79.80 & 89.20 & 98.20 & 57.20 & 65.70 & 94.80 & 89.50 & 66.50 & 67.86 & 70.80 & 77.96 \\
& AUROC & 93.05 & 97.18 & 99.66 & 82.97 & 90.50 & 99.13 & 97.87 & 87.54 & 90.47 & 91.84 & 93.02 \\
\midrule
\multirow{3}{*}{TALL~\cite{xu2023tall}}
& R & 51.20 & 65.20 & 93.40 & 32.00 & 61.60 & 94.80 & 81.80 & 49.20 & 25.00 & 53.60 & 60.78 \\
& BAcc & 75.10 & 82.10 & 96.20 & 65.50 & 80.30 & 96.90 & 90.40 & 74.10 & 61.61 & 76.30 & 79.85 \\
& AUROC & 95.82 & 97.14 & 99.73 & 92.55 & 97.36 & 99.79 & 99.09 & 94.84 & 86.67 & 93.75 & 95.67 \\
\midrule
\multirow{3}{*}{STIL~\cite{gu2021spatiotemporal}}
& R & 73.80 & 70.80 & 43.40 & 1.40 & 2.00 & 45.00 & 13.20 & 7.20 & 1.79 & 11.60 & 27.02 \\
& BAcc & 86.90 & 85.40 & 71.70 & 50.70 & 51.00 & 72.50 & 56.60 & 53.60 & 50.89 & 55.80 & 63.51 \\
& AUROC & 96.43 & 97.77 & 99.34 & 86.66 & 90.56 & 98.88 & 97.04 & 88.16 & 92.57 & 87.52 & 93.49 \\
\midrule
\multirow{3}{*}{NSG-VD~\cite{zhang2025nsgvd}}
& R & 68.33 & 98.33 & 100.00 & 92.50 & 87.50 & 80.00 & 98.33 & 94.17 & 78.57 & 82.50 & 88.02 \\
& BAcc & 81.67 & 98.33 & 96.67 & 91.67 & 90.83 & 88.33 & 95.83 & 94.17 & 88.39 & 88.75 & 91.46 \\
& AUROC & 92.26 & 98.66 & 98.15 & 94.45 & 96.38 & 94.83 & 98.16 & 97.41 & 96.40 & 94.73 & 96.14 \\
\midrule
\multirow{3}{*}{ReStraV$^\dagger$~\cite{interno2026restrav}}
& R & 82.14 & 74.43 & 99.52 & 91.14 & 99.57 & 100.00 & 92.27 & 86.00 & 94.64 & 88.98 & 90.87 \\
& BAcc & 87.47 & 83.61 & 96.16 & 91.97 & 96.18 & 96.39 & 92.53 & 89.39 & 93.72 & 90.88 & 91.83 \\
& AUROC & 93.17 & 92.44 & 99.93 & 96.66 & 99.88 & 100.00 & 97.59 & 95.17 & 98.92 & 96.71 & 97.05 \\
\midrule
\multirow{3}{*}{D3$^{\dagger}$~\cite{zheng2025d3}}
& R & 63.29 & 63.00 & 81.15 & 64.00 & 75.00 & 87.90 & 81.33 & 72.43 & 66.07 & 53.83 & 70.80 \\
& BAcc & 72.13 & 71.99 & 81.07 & 72.49 & 77.99 & 84.44 & 81.16 & 76.70 & 73.53 & 67.41 & 75.89 \\
& AUROC & 79.49 & 80.89 & 88.77 & 81.25 & 86.21 & 91.25 & 89.04 & 84.37 & 80.67 & 77.15 & 83.91 \\
\midrule
\multirow{3}{*}{MM-Det$^\dagger$~\cite{song2024mmdet}}
& R & 93.14 & 95.43 & 97.76 & 46.29 & 40.57 & 98.26 & 93.42 & 68.57 & 55.36 & 45.02 & 73.38 \\
& BAcc & 91.59 & 92.73 & 93.90 & 68.16 & 65.31 & 94.12 & 91.69 & 79.31 & 72.70 & 67.53 & 81.70 \\
& AUROC & 97.40 & 98.10 & 98.42 & 78.45 & 75.23 & 98.89 & 97.14 & 87.12 & 76.34 & 76.22 & 88.33 \\
\midrule
\multirow{3}{*}{\textbf{Ours}}
& R & 95.86 & 96.71 & 97.28 & 93.00 & 95.43 & 99.78 & 99.00 & 95.14 & 83.93 & 68.01 & 92.41 \\
& BAcc & 84.97 & 85.40 & 85.69 & 83.55 & 84.76 & 86.94 & 86.54 & 84.62 & 79.01 & 71.05 & 83.25 \\
& AUROC & 95.65 & 95.76 & 96.63 & 93.15 & 95.11 & 98.71 & 98.27 & 94.96 & 90.40 & 80.09 & 93.87 \\
\bottomrule
\end{tabular*}
}
\vspace{-3mm}
\end{table*}

\section{Experiments}

\paragraph{Datasets and Protocols.}
We evaluate MAST on cross-generator AI-generated video benchmarks.
On GenVideo~\cite{chen2024demamba}, we train on $10{,}000$ real clips
from Kinetics-400~\cite{kay2017kinetics} and $10{,}000$ generated clips
from Pika~\cite{pika2024}. We test on ten unseen sources: ModelScope,
MorphStudio, MoonValley, HotShot, Show1, Gen2, Crafter, Lavie, Sora,
and WildScrape~\cite{wang2023modelscope,morphstudio,moonvalley,hotshot,zhang2025show,gen2,chen2023videocrafter1,wang2025lavie,sora}.
For the backbone-matched comparison, we use IvyFake~\cite{jiang2025ivyfake}
with three training generators (Latte~\cite{ma2024latte},
Pika~\cite{pika2024}, OpenSora~\cite{opensora2024}) and nine unseen
test generators (Crafter, Gen2, HotShot, Lavie, ModelScope, MoonValley,
MorphStudio, Sora, WildScrape). Ablation studies use GenVideo.
Because the GenVideo test set contains far more real than generated
clips, we report balanced accuracy (BAcc), the mean of fake recall (R)
and real-video specificity, together with R and AUROC. On IvyFake, we
report ACC and AUROC over both classes. We average metrics across unseen generators and, following NSG-VD~\cite{zhang2025nsgvd}, select the model with the highest validation AUROC.

\subsection{Main Results}
\label{sec:main_results}

\paragraph{A) Cross-Generator Results.}
Table~\ref{tab:genvideo_pika} compares MAST with eight ANN-based
detectors~\cite{chen2024demamba,tan2024npr,xu2023tall,gu2021spatiotemporal,song2024mmdet,zhang2025nsgvd,interno2026restrav,zheng2025d3}.
MAST achieves $92.41\%$ mean R, $83.25\%$ mean BAcc, and $93.87\%$ mean
AUROC, comparable in AUROC to DeMamba ($93.88\%$), STIL ($93.49\%$),
and NPR ($93.02\%$), while training only a $9.3$\,M-parameter spiking
temporal module on a frozen backbone. On HotShot, Show1, and Lavie,
MAST exceeds all three in AUROC by at least $3.6$ points. Selecting by mean fake recall instead gives $97.14\%$ mean R at $92.71\%$ mean AUROC.
Published and reevaluated results retain their respective test
populations and operating points.

\begin{figure}[h]
\centering

\includegraphics[width=.90\linewidth]{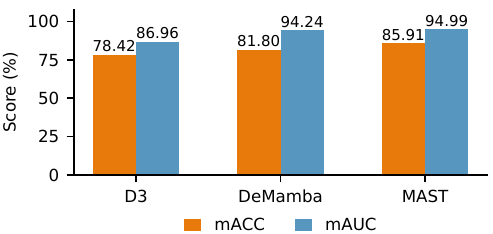}
\caption{IvyFake comparison with the same X-CLIP-B/16 backbone.
Temporal modules: D3 (second-order features), DeMamba (Mamba),
and MAST (SNN).}

\vspace{-5mm}
\label{fig:ivyfake_backbone}

\end{figure}

\begin{figure}[H]
    \centering
    \vspace{-2mm}
    \includegraphics[width=0.9\columnwidth]{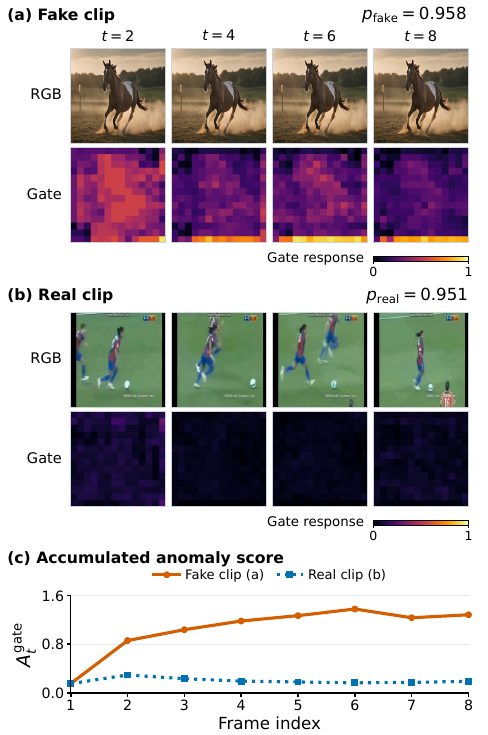}
    \vspace{-3mm}
    \caption{The fake clip (a) shows a stronger lower-border response
    than the real clip (b) and a higher anomaly score from frame~2 (c).}
    \label{fig:case_analysis}
\end{figure}

\paragraph{B) Backbone-Matched Results.}
\label{sec:ivyfake_compare}
To isolate the temporal module, we fix the X-CLIP-B/16 backbone and
compare D3~\cite{zheng2025d3}, DeMamba~\cite{chen2024demamba,gu2023mamba},
and MAST under the multi-generator training setting of
IvyFake~\cite{jiang2025ivyfake} (Figure~\ref{fig:ivyfake_backbone}).
Both learned temporal modules outperform D3's fixed second-order
descriptor, and MAST is the strongest of the three. Its margin over
DeMamba is small in mAUC and larger in mACC, indicating that the two
rank clips similarly while MAST separates them better at the decision
threshold.
\vspace{-4mm}
\paragraph{C) Qualitative Analysis.}
Figure~\ref{fig:case_analysis} traces the gate response and anomaly
score for one fake and one real clip. In the fake clip, the gate
response concentrates along the lower border from $t{=}4$ and the
anomaly score rises, matching the border-localized pattern in
Figure~\ref{fig:teaser}(a); in the real clip, the gate stays nearly
silent and the score remains flat.

\begin{table*}[t!]
\centering
\begin{minipage}[t]{0.48\linewidth}
  \centering
  \caption{Architecture composition ablation. Each row uses a different
  subset of MAST. All metrics are generator-wise means (\%).}
  \label{tab:ablation_branch}

  {\footnotesize
  \setlength{\tabcolsep}{4pt}
  \renewcommand{\arraystretch}{1.0}
  \begin{tabular}{lccc}
    \toprule
    \textbf{Configuration} & \textbf{R} & \textbf{mBACC} & \textbf{mAUC} \\
    \midrule
    \textbf{Full MAST (ours)}    & \textbf{92.41} & \textbf{83.25} & \textbf{93.87} \\
    \midrule
    Semantic Trajectory Branch only                  & 85.79 & 82.34 & 91.16 \\
    SDTB only                 & 77.83 & 81.35 & 89.72 \\
    MAST w/o pseudo-event (RGB only)  & 81.82 & 77.71 & 85.56 \\
    MAST w/o curvature                & 89.49 & 80.59 & 92.24 \\
    \bottomrule
  \end{tabular}
  }
\end{minipage}\hfill
\begin{minipage}[t]{0.48\linewidth}
  \centering
  \caption{Pseudo-event channel leave-one-out ablation. Check marks
  indicate channels kept. All metrics are generator-wise means (\%).}
  \label{tab:ablation_channel}

  {\footnotesize
  \setlength{\tabcolsep}{3pt}
  \renewcommand{\arraystretch}{1.0}
 \begin{tabular}{ccccc|ccc}
\toprule
HF & Sobel & AbsDiff & Diff2 & Chroma & R & mBACC & mAUC \\
\midrule
\checkmark & \checkmark & \checkmark & \checkmark & \checkmark & \textbf{92.41} & \textbf{83.25} & \textbf{93.87} \\
\midrule
           & \checkmark & \checkmark & \checkmark & \checkmark & 90.28 & 81.89 & 91.42 \\
\checkmark &            & \checkmark & \checkmark & \checkmark & 90.66 & 81.56 & 92.28 \\
\checkmark & \checkmark &            & \checkmark & \checkmark & 90.97 & 82.14 & 92.23 \\
\checkmark & \checkmark & \checkmark &            & \checkmark & 89.16 & 81.95 & 91.10 \\
\bottomrule
\end{tabular}
  }
\end{minipage}
\end{table*}

\subsection{Ablation Studies}
\label{sec:ablation}

\paragraph{A+B) Architecture and Channels.}
Full MAST outperforms either branch alone, suggesting that the temporal
and semantic branches provide complementary features
(Table~\ref{tab:ablation_branch}). With raw RGB input, the fused model
falls below the semantic branch alone, so the temporal branch helps
only when its input is cast as pseudo-events. Removing curvature also
lowers performance, suggesting that changes in the direction of patch
embeddings are useful alongside displacement magnitude. Each
pixel-level channel contributes, and removing Diff2 yields the lowest mAUC even though AbsDiff is retained
(Table~\ref{tab:ablation_channel}), suggesting that the SDTB benefits
from explicitly encoding how local frame differences change across
consecutive intervals.

\begin{table}[h!]
\centering
\caption{ANN vs.\ SNN temporal modules with RGB or pseudo-event
inputs. R and mAUC (\%): mean $\pm$ std over three seeds. mJ: module
energy per clip, estimated from operation counts in 45\,nm
CMOS~\cite{yao2025sdtv3}; ANN energy is seed-independent.}
\label{tab:ablation_annsnn}

{\footnotesize
\setlength{\tabcolsep}{4pt}
\renewcommand{\arraystretch}{1.05}
\begin{tabular}{@{}lccc@{}}
\toprule
\textbf{Module + input} & \textbf{R} & \textbf{mAUC} & \textbf{Energy (mJ)} \\
\midrule
ANN + RGB           & $82.17_{\pm0.32}$ & $90.19_{\pm0.35}$ & $85.6$ \\
ANN + pseudo-events & $88.81_{\pm0.48}$ & $91.73_{\pm0.33}$ & $85.6$ \\
\midrule
SNN + RGB           & $81.82_{\pm0.78}$ & $85.56_{\pm0.82}$ & $1.34_{\pm0.43}$ \\
\textbf{SNN + pseudo-events}
                    & $92.41_{\pm0.75}$ & $93.87_{\pm0.78}$ & $1.24_{\pm0.15}$ \\
\bottomrule
\end{tabular}
}
\vspace{-4mm}
\end{table}

\paragraph{C) SNN vs.\ ANN.}
We replace the SDTB with an ANN Transformer of equal depth and width
($L{=}8$, $d{=}256$; $10.7$\,M vs.\ $9.3$\,M parameters;
Appendix~\ref{app:snn_vs_ann}). Pseudo-events raise mAUC for both
modules, but far more for the SNN: with raw RGB the SNN trails the
ANN, whereas with pseudo-events it surpasses it
(Table~\ref{tab:ablation_annsnn}). Under the standard operation-count
energy estimate~\cite{yao2025sdtv3}, the spiking module is also far
cheaper, although the frozen backbone dominates the end-to-end cost.

\paragraph{D) Learnable LIF Parameters.}

\begin{figure}[h!]
\centering

\includegraphics[width=0.9\linewidth]{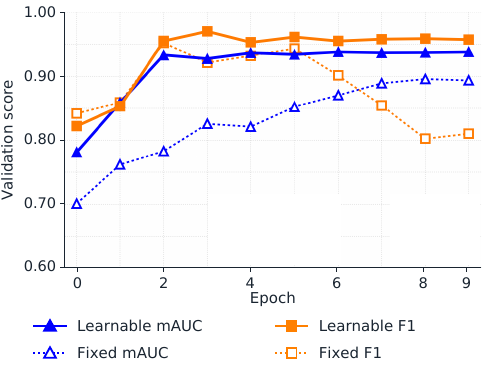}
\caption{Training dynamics of learnable vs.\ fixed LIF.
Fixed: $\tau{=}2.0$, $V_{\text{th}}{=}1.0$.}
\label{fig:tau_dynamics}

\end{figure}

We compare learnable per-channel $\tau_c$ and threshold against a
fixed LIF baseline ($\tau{=}2.0$, $V_{\text{th}}{=}1.0$).
Learnable LIF reaches $93.7\%$ mAUC and $95.7\%$ F1 by epoch 2
(Figure~\ref{fig:tau_dynamics}), while fixed LIF has lower
validation scores and declining F1 after epoch 5. These results
support jointly learning channel-wise time constants and firing thresholds.

\section{Conclusion}
\label{sec:conclusion}

MAST represents temporal forensic evidence as multi-channel pseudo-events
and integrates it with per-channel learnable spiking timescales alongside
frozen semantic features. Trained on a single generator, it reaches
$93.87\%$ mean AUROC and $92.41\%$ mean fake recall on ten unseen GenVideo
sources. Ablations indicate that pseudo-events substantially improve the
spiking branch over raw RGB input, and that the fused model improves
over either pathway alone. With suitable input preprocessing, SNNs can
thus exploit their strength in temporal processing and become
comparable to an ANN of equal depth and width.

\noindent\textbf{Limitations and future work.}
SNNs still trail ANNs in raw performance: MAST remains below the
strongest ANN-based detectors and reaches only $76.81\%$ overall
accuracy on GenVidBench Main
(Appendix~\ref{app:genvidbench_main}), failing mainly on the
image-to-video generators MuseV~\cite{musev} and
SVD~\cite{blattmann2023stable}; short source clips may limit the
temporal evidence available to eight-frame sampling. Extending the
hybrid design to event-native data, spiking encoders, and temporal
localization is a promising direction, and whether the gate maps reflect
generation artifacts rather than dataset-specific cues remains to be
tested.

\section*{LLM Usage Statement}
The authors used large language model tools only in a limited supporting role. The manuscript draft was written by the authors, and OpenAI ChatGPT and Anthropic Claude Sonnet 5.5 were used only for grammatical correction and improving the naturalness of the writing. OpenAI GPT 6 Astra and Anthropic Claude Opus 5.5 were used to assist with code generation and figure visualization, and all generated outputs were reviewed and verified by the authors. The full manuscript or substantial portions of it were also provided to OpenAI ChatGPT and Anthropic Claude to obtain reviewer style feedback. No LLM or AI tool was used for related work and citations, experiment design, experiment execution, result analysis, data generation or annotation, or for generating the core research ideas, hypotheses, claims, or conclusions. All methods, experiments, analyses, and conclusions were designed, performed, and validated by the authors.

{
    \small
    \bibliographystyle{ieeenat_fullname}
    \bibliography{references}
}

\clearpage
\onecolumn
\appendix
\begin{center}
  {\Large\bfseries Appendix\par}
\end{center}
\vspace{0.5em}
\renewcommand{\topfraction}{0.92}
\renewcommand{\textfraction}{0.06}
\renewcommand{\floatpagefraction}{0.8}

\section{Notation and Metrics}
\label{app:eq_details}

We define the pixel-level and feature-space statistics reported in
Section~\ref{sec:structure}.

\subsection{Hoyer Sparsity}
The Hoyer sparsity~\cite{hoyer2004non} of a tensor $x \in
\mathbb{R}^N$ with $N$ entries is a unit-invariant measure of how
concentrated the signal is in space, normalized to $[0,1]$:
\begin{equation}
\mathrm{Hoyer}(x) \;=\; \frac{\sqrt{N} - \|x\|_1 / \|x\|_2}{\sqrt{N} - 1}.
\label{eq:hoyer}
\end{equation}
Values close to $1$ correspond to spike-like distributions
concentrated at a few entries; values close to $0$ correspond to
uniform distributions. Section~\ref{sec:structure} reports
$\mathcal{S}_\mathrm{rgb}$ and $\mathcal{S}_\mathrm{res}$ as the
clip-averaged Hoyer sparsity of the raw RGB image and of the
high-frequency Laplacian temporal residual respectively, with
$\mathcal{S}_\mathrm{res} = \tfrac{1}{T-1}\sum_{t=2}^{T}
\mathrm{Hoyer}(\Delta^{\mathrm{HF}}_t)$.

\subsection{Spectral Centroid}
For each clip we form the per-frame spatial mean
$s(t) = \mathrm{spatial\_mean}(\Delta^{\mathrm{HF}}_t)$ of the
high-frequency Laplacian residual and compute its discrete-Fourier
power spectrum and spectral centroid:
\begin{equation}
P(f) \;=\; \big|\mathrm{FFT}(s)\big|^2,
\qquad
f_c \;=\; \frac{\sum_f f \cdot P(f)}{\sum_f P(f)}.
\label{eq:fc}
\end{equation}
The centroid is the energy-weighted mean temporal frequency.
Frequencies are normalized by the frame rate, so $f$ is in cycles per
frame and lies in $[0, 0.5]$. Main-text Figure~2(c) plots
$\log_{10} f_c$, which is why its horizontal axis is negative.

\subsection{Trajectory Curvature}
Given the X-CLIP frame embeddings $Z_t$ and their per-step
displacement $\Delta Z_t = Z_t - Z_{t-1}$, the angular
curvature at frame $t$ is the angle between two consecutive
displacement vectors,
\begin{equation}
\theta_t \;=\; \arccos\!\frac{
  \big\langle \Delta Z_{t},\, \Delta Z_{t-1} \big\rangle}{
  \big\| \Delta Z_{t} \big\|_2 \, \big\| \Delta Z_{t-1} \big\|_2}.
\label{eq:curv_video}
\end{equation}
Low $\theta_t$ corresponds to a smooth latent path, high $\theta_t$
to sharp directional jumps. Section~\ref{sec:structure} reports the
clip-median $\theta$ as a per-clip statistic.

\subsection{Trajectory Volume}
\label{app:vol}
The volume statistic \textit{Vol.}\ in
Table~\ref{tab:per_gen_stats} summarizes the spread of each
trajectory in latent space. We project the X-CLIP frame
embeddings $Z_t \in \mathbb{R}^{D}$ to their first
three PCA components, $z(t) = U^\top Z_t \in \mathbb{R}^3$, where
$U$ is fit jointly on the union of all real and generated clips in
the analyzed pool. For a video of length $T$, we compute the volume of the
convex hull formed by the projected trajectory points:
\begin{equation}
\mathrm{Vol.} \;=\; \mathrm{Volume}\big(\mathrm{Conv}(\{z(t)\}_{t=1}^T)\big).
\label{eq:vol}
\end{equation}
For this analysis, each clip contributes $T=16$ frame embeddings to its convex hull. The PCA projection is used only for this volume summary and for trajectory visualization; it is \emph{not} the feature space in which the classifier operates.

\paragraph{Trajectory statistics.}
\label{app:trajectory_summary}
The angular curvature in Eq.~\eqref{eq:curv_video} and the
distance-based summaries below are computed directly in the full
$D$-dimensional X-CLIP feature space, without dimensionality
reduction. Given frame embeddings $\{Z_t\}_{t=1}^{T}$,
we form first- and second-order temporal differences,
\begin{equation}
d^{(1)}_t \;=\; \big\| Z_{t+1} - Z_{t} \big\|_2,
\qquad
d^{(2)}_t \;=\; d^{(1)}_{t+1} - d^{(1)}_{t},
\end{equation}
and report $\{\mathrm{mean}, \mathrm{std}, \mathrm{max}, \mathrm{min}\}$
of each sequence, together with the analogous cosine-distance
variant. These summaries form $F_{\mathrm{traj}}$ in the classifier fusion
of Section~\ref{sec:fusion}; they are distinct from the two patch-level
trajectory channels supplied to the SDTB. Per-frame features are $L_2$-normalized before the
cosine-distance computation; no other normalization is applied.
The PCA projection used for the convex-hull volume above is
\emph{not} applied here.

\section{Residual Channels}
\label{app:channels}

The SDTB uses five pixel residuals (HF, Sobel, Chroma, AbsDiff, and
Diff2) and two patch-trajectory channels (displacement and curvature),
seven channels in total. The main text illustrates four of the pixel
channels (main-text Figures~3 and~6). The Chroma channel is documented
here: its definition is given in Appendix~\ref{app:chroma_neg}, its
per-generator distribution in Figure~\ref{fig:app_chroma}, and example
residual maps in Figures~\ref{fig:gallery_gen2}
and~\ref{fig:gallery_show1}. Appendix~\ref{app:per_generator}
compares the distributions of the pixel residuals on IvyFake.

\subsection{HF: Laplacian Difference}
The HF channel measures changes in the spatial Laplacian between adjacent frames.
\begin{align}
  Y_t       &= 0.299\, R_t + 0.587\, G_t + 0.114\, B_t \\
  L_t       &= \mathrm{Conv2d}(Y_t;\, K_\mathrm{Lap}),\quad
               K_\mathrm{Lap} = \begin{bmatrix} 0 & 1 & 0 \\ 1 & -4 & 1 \\ 0 & 1 & 0 \end{bmatrix} \\
  \Delta^\mathrm{HF}_t &= |L_t - L_{t-1}|,\quad t = 2, \dots, T.
\end{align}

\begin{figure*}[tp]
  \centering
  \bigfig{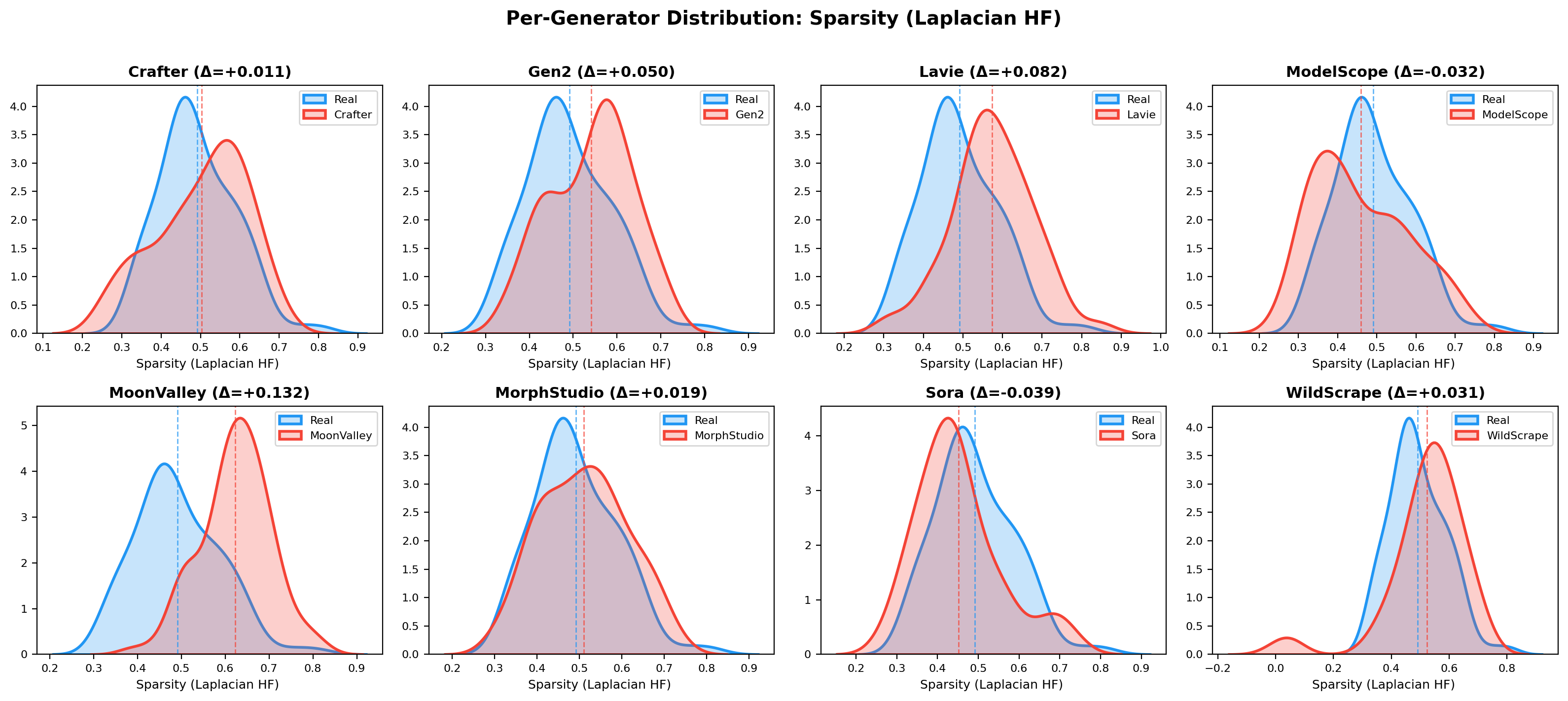}
  \caption{Per-generator Laplacian HF sparsity distributions on IvyFake.}
  \label{fig:app_hf}
\end{figure*}

\subsection{Sobel: Edge Difference}
The Sobel channel measures changes in edge magnitude between adjacent frames.
\begin{align}
  G^x_t &= \mathrm{Conv2d}(Y_t;\, K_\mathrm{Sobel\text{-}x}),\quad
  G^y_t = \mathrm{Conv2d}(Y_t;\, K_\mathrm{Sobel\text{-}y}) \\
  M_t &= \sqrt{(G^x_t)^2 + (G^y_t)^2 + \epsilon} \\
  \Delta^\mathrm{Sobel}_t &= |M_t - M_{t-1}|,
\end{align}
where $K_\mathrm{Sobel\text{-}x}, K_\mathrm{Sobel\text{-}y}$ are the standard
$3 \times 3$ Sobel kernels and $\epsilon = 10^{-6}$.

\begin{figure*}[tp]
  \centering
  \bigfig{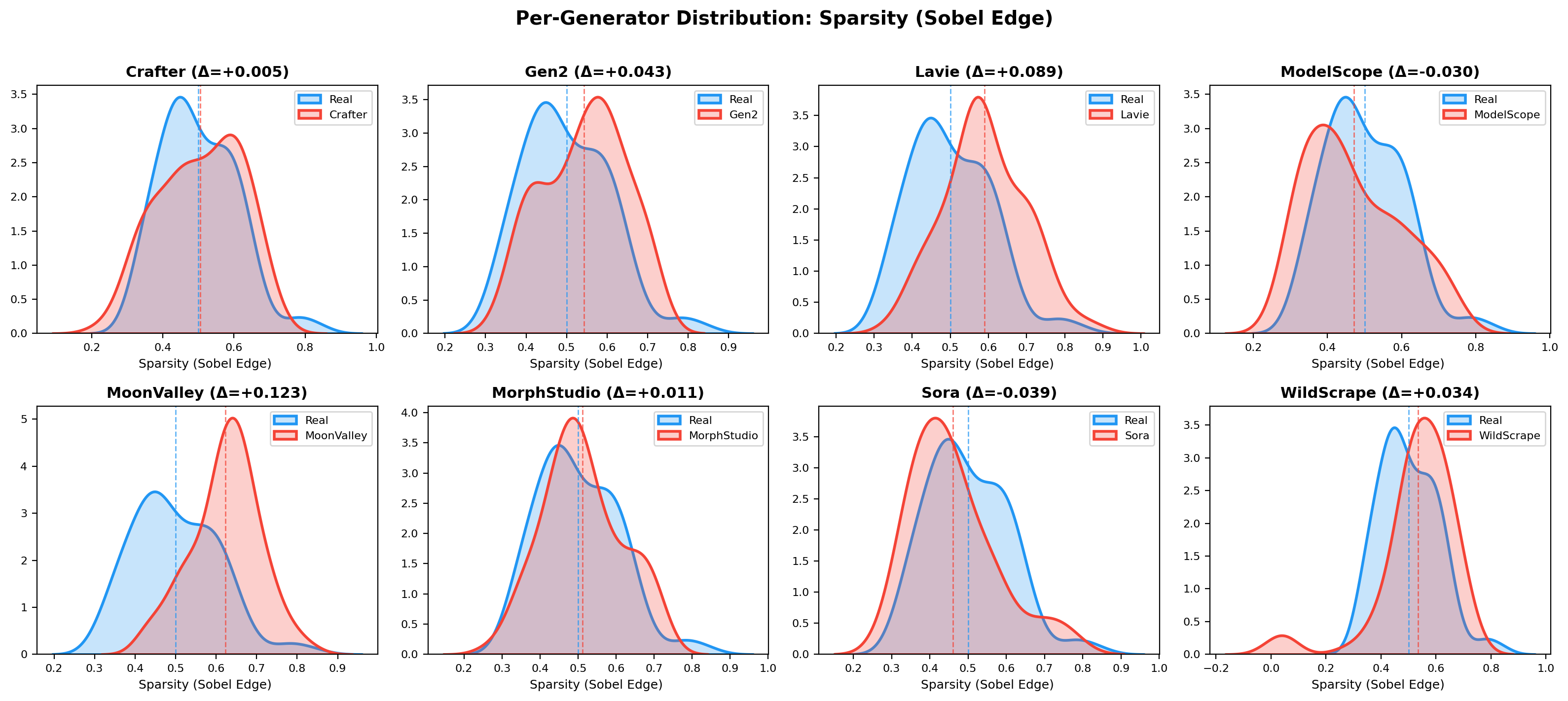}
  \caption{Per-generator Sobel sparsity distributions on IvyFake.}
  \label{fig:app_sobel}
\end{figure*}

\subsection{Chroma: Chromatic Difference}
\label{app:chroma_neg}
\begin{figure*}[tp]
  \centering
  \bigfig{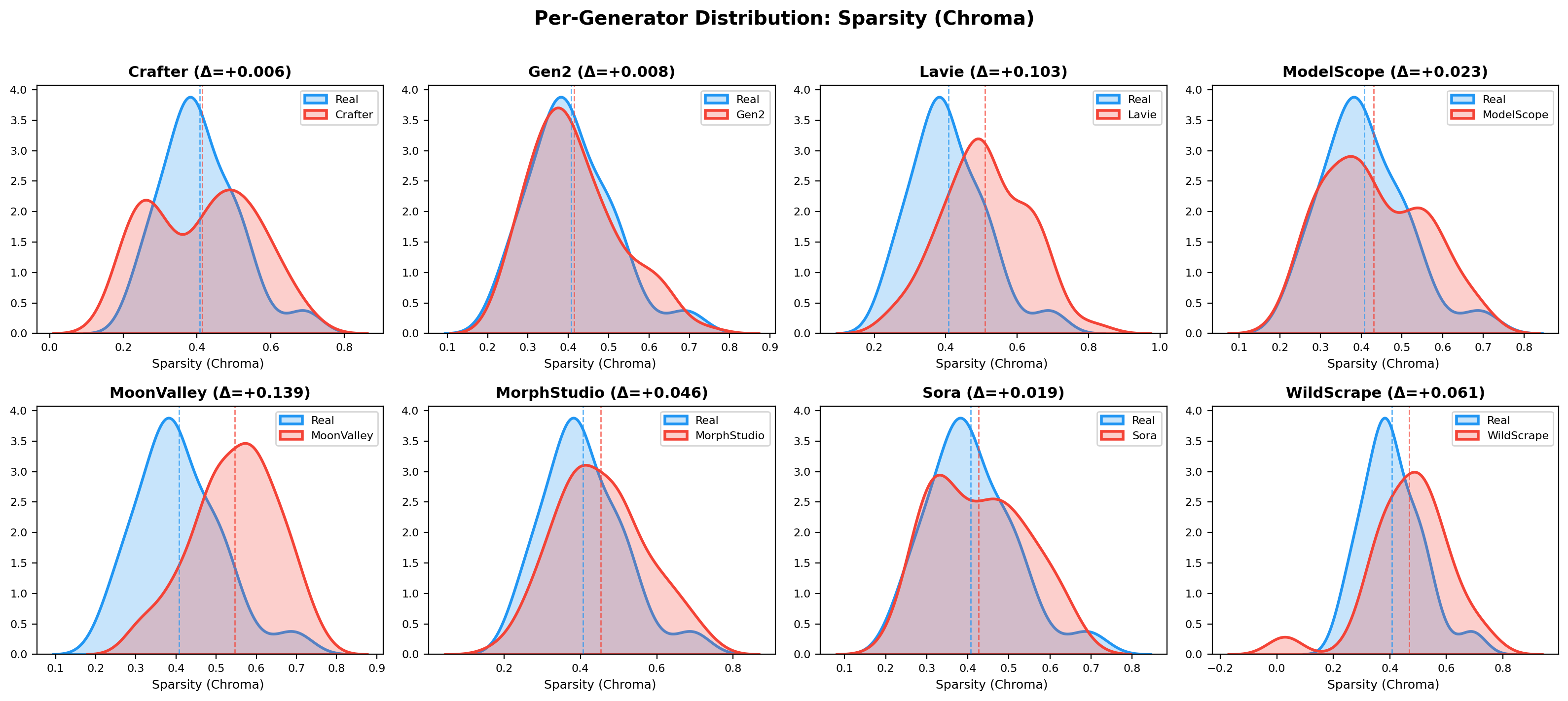}
  \caption{Per-generator chroma sparsity distributions on IvyFake.}
  \label{fig:app_chroma}
\end{figure*}

The Chroma channel measures inter-frame changes in the YCbCr chroma
components, which the luma-based HF and Sobel residuals do not
capture. The channel is defined as
\begin{align}
  Cb_t &= -0.169\, R_t - 0.331\, G_t + 0.500\, B_t + 0.5 \\
  Cr_t &= \phantom{-}0.500\, R_t - 0.419\, G_t - 0.081\, B_t + 0.5 \\
  \Delta^\mathrm{Chroma}_t &= |Cb_t - Cb_{t-1}| + |Cr_t - Cr_{t-1}|.
\end{align}
Figure~\ref{fig:app_chroma} shows its per-generator sparsity
distributions. Chroma is kept in every row of main-text
Table~\ref{tab:ablation_channel}; we do not ablate it individually.

\subsection{AbsDiff: Frame Difference}
The simplest temporal residual: per-pixel absolute RGB difference, averaged
across color channels.
\begin{equation}
  \Delta^\mathrm{abs}_t
  = \frac{1}{3}\sum_{c \in \{R,G,B\}} |F^c_t - F^c_{t-1}|.
\end{equation}

\begin{figure*}[tp]
  \centering
  \bigfig{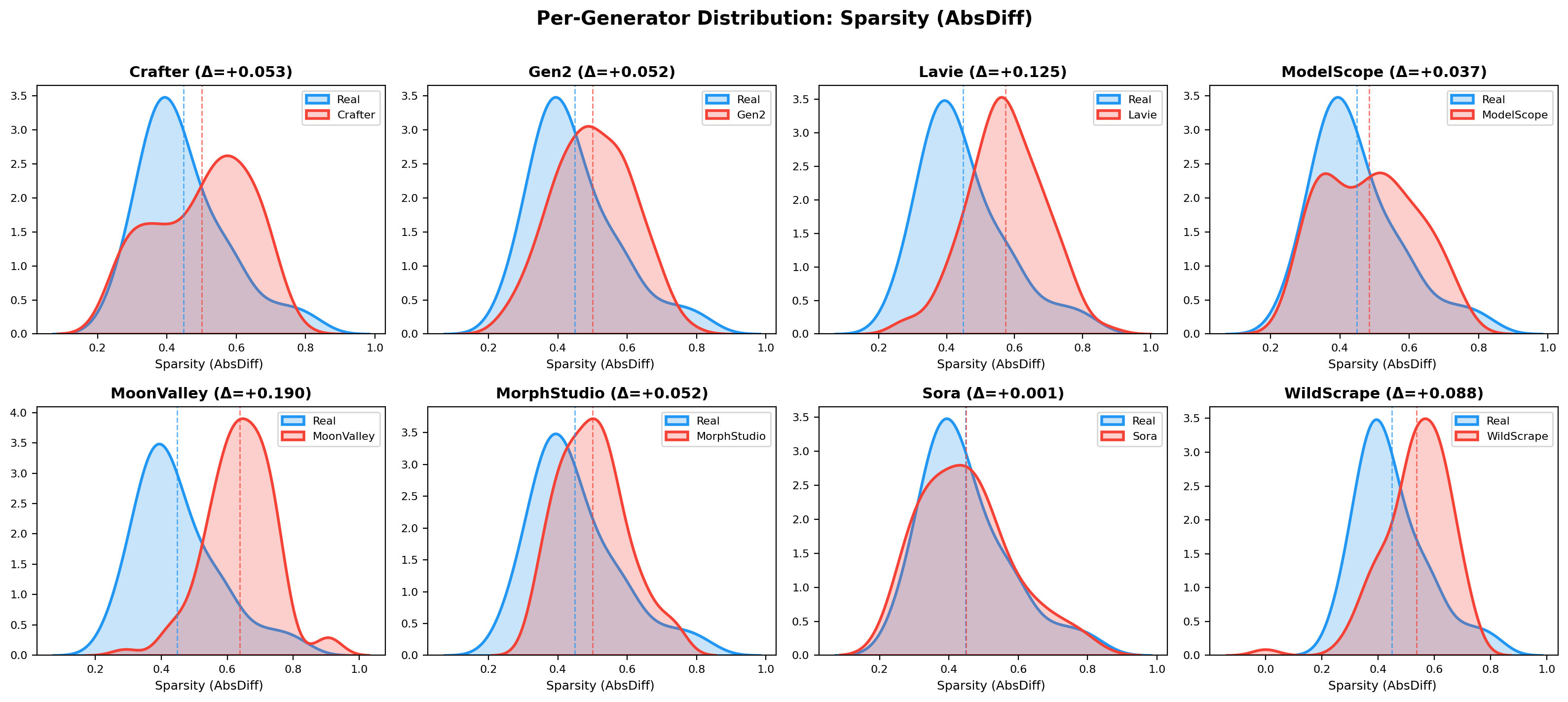}
  \caption{Per-generator absolute difference sparsity distributions on IvyFake.}
  \label{fig:app_absdiff}
\end{figure*}

\subsection{Diff2: Second-Order Difference}
Diff2 measures the change in the magnitude of consecutive AbsDiff residuals.
\begin{equation}
  \Delta^\mathrm{2nd}_t
  = |\Delta^\mathrm{abs}_t - \Delta^\mathrm{abs}_{t-1}|,\quad t \geq 3.
\end{equation}

\begin{figure*}[tp]
  \centering
  \bigfig{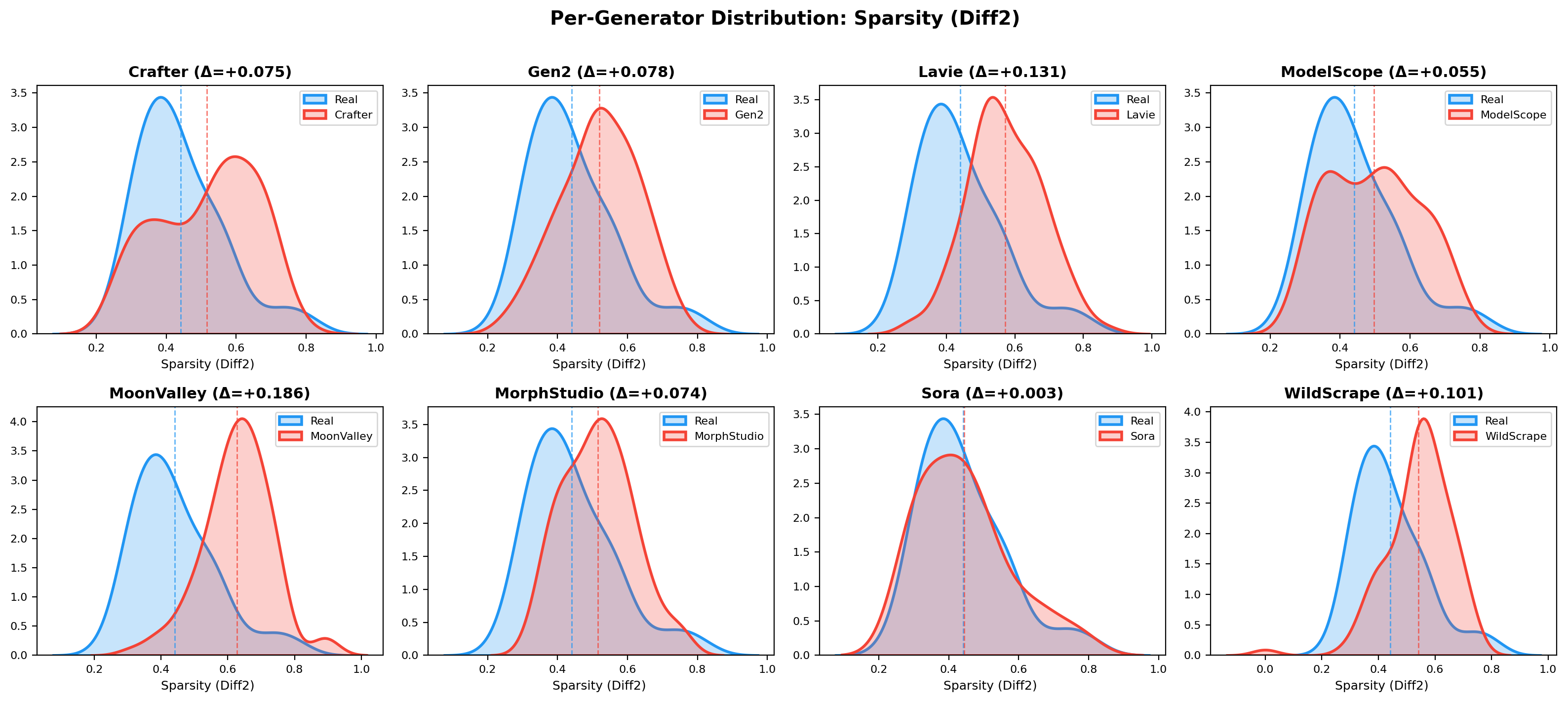}
  \caption{Per-generator second-order difference sparsity distributions on IvyFake.}
  \label{fig:app_diff2}
\end{figure*}

\subsection{Residual Post-processing}
\label{app:postproc}
Each pixel residual map $x_t$ is computed on raw RGB frames
(Section~\ref{sec:method}). Before entering the SDTB, it is divided
by a per-frame mean $\bar{x}_t$, pooled to the $14 \times 14$ X-CLIP
patch grid, multiplied by a learnable scale $\mathrm{softplus}(w)$,
and passed through the soft threshold of
Eq.~\eqref{eq:pseudo_event} with $c_{\mathrm{th}} = 0.1$. HF, Sobel,
and Chroma share one mean, taken over the three maps of the frame,
and one scale; AbsDiff and Diff2 each use their own spatial mean and
scale. The first frame, which has no temporal predecessor, is set to
zero:
\begin{align}
  \tilde{x}_t &= x_t / \big(\bar{x}_t + \epsilon\big) \\
  x'_t &= \mathrm{AdaptiveAvgPool2d}(\tilde{x}_t,\, 14 \times 14) \\
  x''_t &= \mathrm{softplus}(w) \cdot x'_t,\quad
  x'''_t = \sigma\!\left((x''_t - c_{\mathrm{th}})/\beta\right) \\
  x'''_1 &\leftarrow 0.
\end{align}
After this rescaling, the pseudo-events encode where changes
concentrate within a frame rather than their absolute magnitude.
The five channels are concatenated with the two trajectory channels
(Section~\ref{sec:method}), which are rescaled by their joint
per-frame mean and are not thresholded, to form the seven-channel
input to the SDTB.

\section{IvyFake Distributions}
\label{app:per_generator}

This appendix reports per-generator distributions on
IvyFake~\cite{jiang2025ivyfake} for the five residual channels defined
in Appendix~\ref{app:channels} and for the trajectory curvature
referenced in Section~\ref{sec:structure}, together with a
t-SNE visualization of the feature spaces produced by the two
branches and by their fusion. The aim is to (i) examine whether the
real-vs-fake gaps observed in the GenVidBench Pair-1 analysis
recur on the broader, multi-generator IvyFake test pool, and
(ii) characterize how each residual cue behaves across generators.

\subsection{Channel Sparsity}
\label{app:distributions}

We report per-generator Hoyer sparsity distributions for the five
residual channels (HF, Sobel, Chroma, AbsDiff, Diff2). For readability,
the figures are placed inline with their channel definitions in
Appendix~\ref{app:channels}: Figure~\ref{fig:app_hf} (HF),
Figure~\ref{fig:app_sobel} (Sobel), Figure~\ref{fig:app_chroma}
(Chroma), Figure~\ref{fig:app_absdiff} (AbsDiff), and
Figure~\ref{fig:app_diff2} (Diff2).

The overlap with the real reference varies across generators and channels.
The HF and Sobel maps emphasize changes in local spatial structure,
Chroma captures changes in color, and AbsDiff and Diff2 summarize
changes in pixel values and in successive residual magnitudes.
Main-text Table~\ref{tab:ablation_channel} evaluates the contributions
of the pixel channels to the detector on GenVideo.

\subsection{Trajectory Curvature}

We additionally report trajectory curvature in the X-CLIP feature
space on IvyFake, computed per clip with the displacement-vector
formulation of Appendix~\ref{app:eq_details}. We plot the
per-generator kernel density estimate against the real reference
distribution and report the mean shift
$\Delta = \mu_\mathrm{fake} - \mu_\mathrm{real}$ in the subplot
title.

\begin{figure*}[tp]
  \centering
  \bigfig{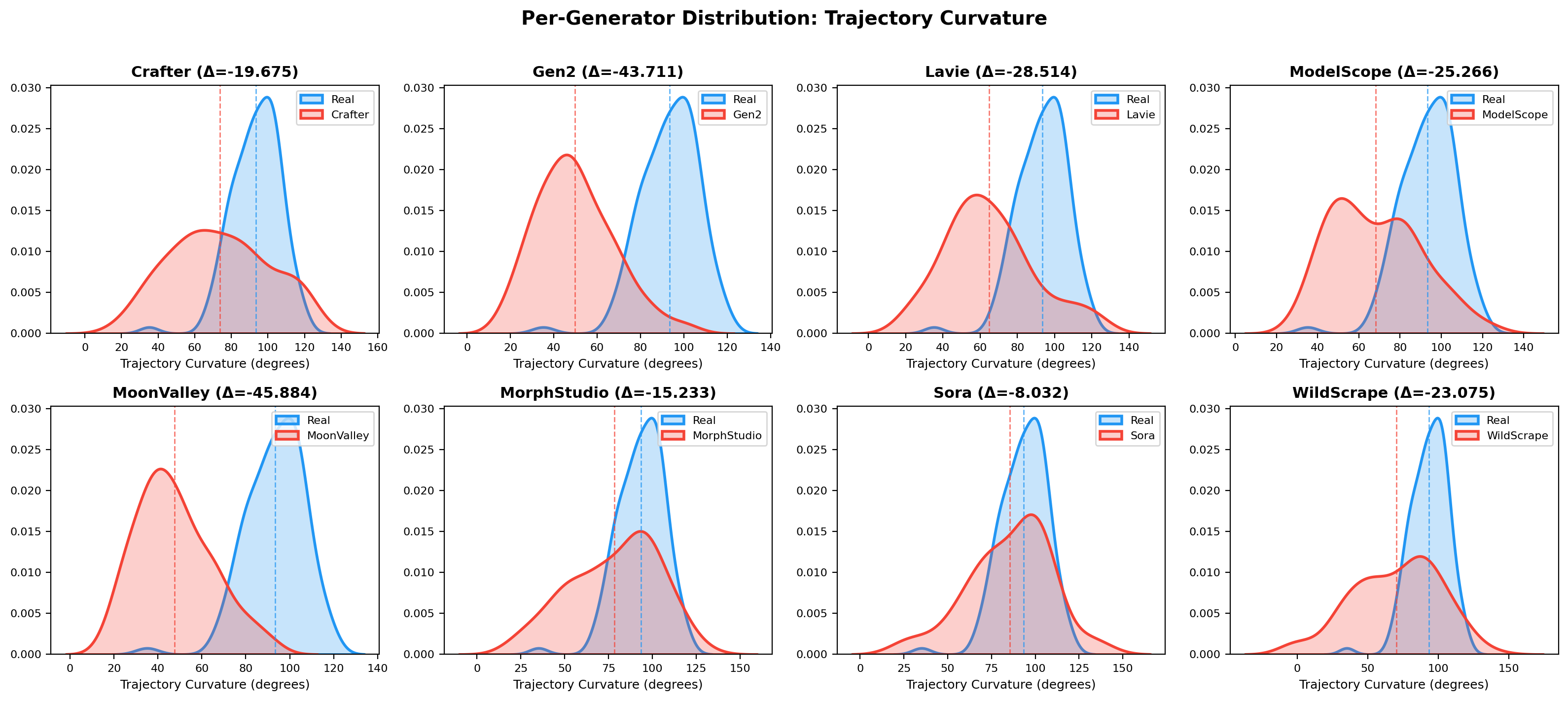}
  \caption{Per-generator trajectory curvature distributions on IvyFake.}
  \label{fig:app_curvature}
\end{figure*}

Trajectory curvature ($\theta$) is lower than the real reference
for all eight generated sources shown in
Figure~\ref{fig:app_curvature}, consistent with the GenVidBench
Pair-1 observation (Section~\ref{sec:structure}).

\subsection{Feature-Space t-SNE}

\begin{figure*}[p]
  \centering
  \bigfig{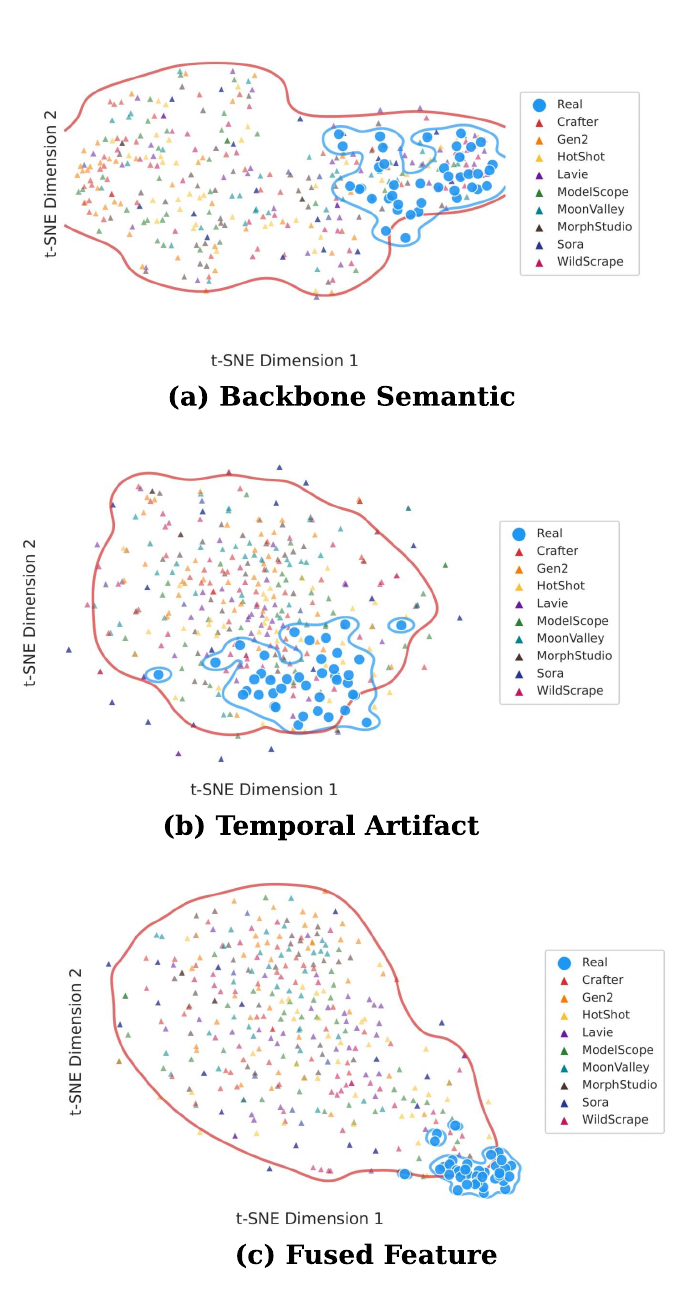}
  \caption{t-SNE projections of MAST feature spaces on IvyFake:
  (a) backbone semantic features, (b) temporal artifact features,
  and (c) fused features. Real clips are blue circles; generated
  clips use colored markers for each source.}

  \label{fig:tsne}
\end{figure*}

Figure~\ref{fig:tsne} shows t-SNE projections of three MAST
representations: (a) the backbone semantic feature
$Z_{\mathrm{video}}$, (b) the temporal artifact feature
$F_{\mathrm{anom}}\parallel F_{\mathrm{gate}}$, and (c) the fused
feature $F_{\mathrm{final}}=[Z_{\mathrm{video}}\parallel
F_{\mathrm{anom}}\parallel F_{\mathrm{gate}}\parallel
F_{\mathrm{traj}}]$. Real clips are blue circles, and generated
clips use colored markers for each source.
These projections qualitatively illustrate the arrangement
of real and generated clips in each feature space.

\section{SNN and ANN Architectures}
\label{app:snn_vs_ann}

\subsection{Architecture Details}
\label{app:arch_details}

For pseudo-event inputs, the ANN and SNN share the interface in
Table~\ref{tab:arch_io}. Both use eight blocks with hidden width 256.
Table~\ref{tab:arch_spec} pairs the operators used at each stage.
Main-text Table~\ref{tab:ablation_annsnn} also evaluates both modules
with RGB inputs.

\begin{table*}[tp]
\centering
\caption{Common input/output interface of the SDTB and the
ANN counterpart.}
\label{tab:arch_io}

{\small
\setlength{\tabcolsep}{6pt}
\begin{tabular}{ll}
  \toprule
  \textbf{Tensor} & \textbf{Shape} \\
  \midrule
  Input event grid          & $\mathbb{R}^{B \times T \times C \times H' \times W'}$ \\
  Channels $C$              & $5$ pixel residual (HF, Sobel, Chroma, AbsDiff, Diff2) $+\,2$ trajectory (displacement, curvature) \\
  Grid $H' \times W'$     & $14 \times 14$ (X-CLIP patch grid) \\
  Time length $T$           & $8$ frames at $\mathrm{fps}{=}8$ \\
  \midrule
  Gate map                  & $G \in [0,1]^{B \times T \times H' \times W'}$ \\
  Token gates               & $\mathbb{R}^{B \times T \times N}$, $N{=}H'W'$ \\
  Anomaly trace             & $\mathcal{A}^{\mathrm{gate}} \in \mathbb{R}^{B \times T}$ \\
  \bottomrule
\end{tabular}
}
\end{table*}

\paragraph{ANN baseline (CNN-Transformer).}
The ANN baseline replaces every spike-driven component with its
standard continuous counterpart while preserving depth, hidden width, head
count, and MLP ratio. The PerChannel preprocessing keeps a
per-channel $3{\times}3$ Conv but drops the LIF state, applying
LayerNorm + GELU per frame instead, removing explicit
time-accumulation in the ANN preprocessing. The block follows a standard post-norm
($\mathrm{LN}(x + f(x))$) structure, and the attention is a standard softmax
multi-head attention on dense floating-point tensors, flattened over
both space and time. The accumulator is a linear EMA over the
gate-map mean.

\begin{table*}[tp]
\centering
\caption{Stage-by-stage specification of the SDTB and the
ANN counterpart. Both branches use hidden width
$d{=}256$, depth $L{=}8$, $h{=}4$ heads, and mlp\_ratio $r{=}2.5$.}
\label{tab:arch_spec}

{\footnotesize
\setlength{\tabcolsep}{3.5pt}
\begin{tabular}{llll}
  \toprule
  \textbf{Stage} & \textbf{Component} & \textbf{SDTB} & \textbf{ANN baseline} \\
  \midrule
  \multirow{2}{*}{Stage 1}
    & Preprocessing
        & PerChannelLIF: learnable $\tau_c$, $V_{\mathrm{th},c}$
        & PerChannel-LayerNorm + GELU (frame-wise) \\
    & Projection
        & Conv$2$d$_{1 \to D_c, 3 \times 3}$ + GroupNorm, $D_c{=}20$
        & Conv$2$d$_{1 \to D_c, 3 \times 3}$ + BN, $D_c{=}20$ \\
  \midrule
  Stage 2
    & Channel fusion to $d{=}256$
        & Conv$1{\times}1$ ($CD_c \to 256$) + BN
        & Conv$1{\times}1$ ($CD_c \to 256$) + BN + GELU \\
  \midrule
  \multirow{4}{*}{Stage 3}
    & Block depth
        & $L{=}8$ Modified SDT-V3 Blocks
        & $L{=}8$ ANN-Blocks (post-norm) \\
    & Local mixing
        & None ($1{\times}1$ convolutions only)
        & ConvStem (single $3{\times}3$ + BN) \\
    & Attention
        & Multispike Linear Attention (E-SDSA), per frame
        & Softmax MHA, $h{=}4$, $d_h{=}64$ \\
    & MLP
        & Linear $\to$ Multispike $\to$ Linear $\to$ Multispike, $r{=}2.5$
        & Linear $\to$ GELU $\to$ Linear, $r{=}2.5$ \\
  \midrule
  Stage 4
    & Per-position gate
        & Conv$1{\times}1$ ($256 \to 1$), $\sigma((\cdot + b_{\mathrm{init}})/\tau_{\mathrm{gate}})$
        & Conv$1{\times}1$ ($256 \to 1$), same $\sigma$ scaling \\
  \midrule
  \multirow{2}{*}{Stage 5}
    & Anomaly accumulator
        & Non-firing LIF: $A[t]{=}(1{-}\tfrac{1}{\tau_{\mathrm{anom}}})A[t{-}1]+\tfrac{1}{\tau_{\mathrm{anom}}}g_t$
        & EMA: $A[t]{=}\alpha A[t{-}1]+(1{-}\alpha)\bar{G}_t$ \\
    & Hyperparameters
        & $\tau_{\mathrm{anom}}{=}2.0$, non-firing ($V_{\mathrm{anom}}{=}\infty$)
        & $\alpha{=}0.5$ (optionally learnable) \\
  \bottomrule
\end{tabular}
}
\end{table*}

\subsection{Operator Differences}
\label{app:operator_diff}

Stage 3 uses E-SDSA adapted from SDT-V3~\cite{yao2025sdtv3}.
Multispike projections of the input $U$ produce the queries, keys, and values:
\begin{equation}
\begin{aligned}
Q_S &= \mathrm{Multispike}(\mathrm{Linear}(U)), \\
K_S &= \mathrm{Multispike}(\mathrm{Linear}(U)), \\
V_S &= \mathrm{Multispike}(\mathrm{Linear}(U)),
\end{aligned}
\end{equation}
\begin{equation}
\mathrm{SpikeAttn}(Q_S,K_S,V_S) \;=\;
\mathrm{Linear}_{out}\Big(\mathrm{Multispike}\big( (Q_S K_S^\top V_S) \cdot \mathrm{scale}\big)\Big).
\label{eq:spikeattn1}
\end{equation}
The discrete $Q_S,K_S,V_S$, each taking $L{+}1$ levels, enter linear
attention without softmax. The ANN instead computes
$\mathrm{Softmax}(QK^\top/\sqrt{d_h})V$ with continuous projections.
The modules also differ in local mixing and temporal accumulation
(Table~\ref{tab:arch_spec}).

\subsection{Parameter Counts}
\label{app:param_breakdown}

The SDTB has $9.3$\,M parameters ($9.03$\,M in the projection, blocks,
and gate, and $9.28$\,M with the feature projections that produce
$F_{\mathrm{anom}}$ and $F_{\mathrm{gate}}$); the main text reports
$10.7$\,M for the ANN counterpart. Both counts cover the temporal
module only; the X-CLIP encoder is frozen in both configurations.
Main-text Table~\ref{tab:ablation_annsnn} reports detection results for
both temporal modules under RGB and pseudo-event inputs.

\subsection{Energy Estimate}
\label{app:energy}

The energy column of main-text Table~\ref{tab:ablation_annsnn} follows
the operation-count estimate commonly used for spike-driven
Transformers~\cite{yao2025sdtv3}, with the 45\,nm CMOS constants
of~\cite{Horowitz2014ISSCC}: $E_{\mathrm{MAC}} = 4.6$\,pJ per
multiply--accumulate and $E_{\mathrm{AC}} = 0.9$\,pJ per accumulate.
For the ANN module, the energy is the analytic multiply--accumulate
count of one $T{=}8$ clip multiplied by $E_{\mathrm{MAC}}$; it does not
depend on the training seed. For the SNN module, the synaptic-operation
count is the dense operation count multiplied by the measured firing
rate of the trained module, and the energy is this count multiplied by
$E_{\mathrm{AC}}$; its variation across seeds comes from the firing
rate. Both values cover the temporal module only. Input construction
and the frozen X-CLIP backbone are excluded, and the backbone accounts
for most of the end-to-end cost. The values are estimates for digital
CMOS rather than measurements on neuromorphic hardware.

\section{GenVidBench Main Results}
\label{app:additional_experiments}
\label{app:genvidbench_main}

We additionally evaluate on the GenVidBench~\cite{ni2026genvidbench}
Main task, covering MuseV~\cite{musev}, SVD~\cite{blattmann2023stable},
CogVideo~\cite{hong2022cogvideo}, Mora~\cite{yuan2024mora}, and HD-VG
real videos~\cite{wang2023videofactory}
(Table~\ref{tab:genvidbench_M}). The table also lists the action
recognition baselines reported in~\cite{interno2026restrav}.
MAST obtains $97.80\%$ on CogVideo, $91.18\%$ on Mora, and $97.53\%$
on HD-VG, but falls to $55.55\%$ on MuseV and $41.99\%$ on SVD, for a
mean accuracy of $76.81\%$ over the five sources. MuseV and SVD are image-to-video
sources with the shortest clips in this benchmark: 13 and 14 frames
($1.08$ and $1.4$ seconds), compared with 33 frames ($4.1$ seconds)
for CogVideo and 24 ($2.4$ seconds) for Mora. This pattern is
consistent with the short-clip limitation noted in the main text.

\begin{table*}[tp]
\centering
\caption{GenVidBench~\cite{ni2026genvidbench} Main (M) task accuracy
(\%). Best per-column in bold; second-best underlined. $\dagger$:
trained and evaluated by us. Other results are sourced from~\cite{interno2026restrav}.
For the six action-recognition baselines, Avg.\ is the overall accuracy
reported in the source, which is not the unweighted mean of the five
columns; for the remaining rows, Avg.\ is the unweighted mean of the
five columns.}
\label{tab:genvidbench_M}

{\small
\setlength{\tabcolsep}{8pt}
\renewcommand{\arraystretch}{1.25}
\begin{tabular}{lcccccc}
\toprule
\textbf{Method} & \textbf{MuseV} & \textbf{SVD} & \textbf{CogVideo} & \textbf{Mora} & \textbf{HD-VG} & \textbf{Avg.} \\
\midrule
TSM~\cite{lin2019tsm}                & 70.37 & 54.70 & 78.46 & 70.37 & 96.76 & 76.40 \\
X3D~\cite{feichtenhofer2020x3d}      & \underline{92.39} & 37.27 & 65.72 & 49.60 & 97.51 & 77.09 \\
MViT V2~\cite{li2022mvitv2}          & 76.34 & \textbf{98.29} & 47.50 & \textbf{96.62} & \underline{97.58} & \underline{79.90} \\
SlowFast~\cite{feichtenhofer2019slowfast} & 12.25 & 12.68 & 38.34 & 45.93 & 93.63 & 41.66 \\
I3D~\cite{carreira2017quo}           &  8.15 &  8.29 & 60.11 & 59.24 & 93.99 & 49.23 \\
VideoSwin~\cite{liu2022video}        & 62.29 &  8.01 & 91.82 & 45.83 & \textbf{99.29} & 67.27 \\
\midrule
NSG-VD$^{\dagger}$~\cite{zhang2025nsgvd}         & 69.2 & 61.7 & 75.7 & 55.3 & 91.05 & 70.59 \\
ReStraV~\cite{interno2026restrav}    & \textbf{93.52} & \underline{94.01} & \underline{93.52} & \underline{92.97} & 91.07 & \textbf{93.01} \\
D3$^{\dagger}$~\cite{zheng2025d3}    & 54.43 & 54.59 & 55.87 & 55.74 & 10.50 & 46.23 \\
\textbf{Ours}                        & 55.55 & 41.99 & \textbf{97.80} & 91.18 & 97.53 & 76.81 \\
\bottomrule
\end{tabular}
}
\end{table*}

\section{SNN Training Details}
\label{app:training_details}

This appendix gives the training objective, the per-channel LIF
settings, and the forward algorithm (Algorithm~\ref{alg:mast}).
Optimizer and schedule settings are listed in
Appendix~\ref{app:params}.

\subsection{Training Objective}
\label{app:ablation_loss}

We optimize the trainable components of MAST while keeping the
X-CLIP encoder frozen. The optimized objective combines three task
losses:
\begin{equation}
    \mathcal{L} = \mathcal{L}_{\mathrm{BCE}}(\hat{y}, y)
                + \lambda_1 \mathcal{L}_{\mathrm{BCE}}(\hat{y}_{\mathrm{snn}}, y)
                + \lambda_2 \mathcal{L}_{\mathrm{supcon}}.
    \label{eq:loss_appendix}
\end{equation}
The spike-rate term listed in the main text
(Eq.~\ref{eq:rate_penalty}) is monitored but not optimized
(Appendix~\ref{app:impl_notes}).

Here $y\in\{0,1\}$ is the real/fake label, $\hat{y}\in[0,1]$ is the
sigmoid prediction on the fused representation $F_{\mathrm{final}}$,
and $\hat{y}_{\mathrm{snn}}\in[0,1]$ is the prediction of an
auxiliary classification head that consumes only the spike-driven
temporal features ($F_{\mathrm{anom}}\!\parallel\!F_{\mathrm{gate}}$).
The main and auxiliary classification losses are standard binary
cross-entropy,
\begin{equation}
    \mathcal{L}_{\mathrm{BCE}}(p, y)
    \;=\; -\,y\log p \;-\; (1-y)\log (1-p),
    \label{eq:bce}
\end{equation}
applied to $\hat{y}$ and $\hat{y}_{\mathrm{snn}}$ respectively.
The supervised contrastive
loss~\cite{khosla2020supcon} acts on the $\ell_2$-normalized fused
features $z_i = F_{\mathrm{final},i}/\lVert F_{\mathrm{final},i}\rVert_2$
within a mini-batch of size $B$,
\begin{equation}
    \mathcal{L}_{\mathrm{supcon}}
    \;=\;
    \sum_{i=1}^{B} \frac{-1}{|P(i)|}
    \sum_{p \in P(i)}
    \log \frac{\exp\!\big(z_i^\top z_p / \tau_{\mathrm{con}}\big)}
              {\sum_{a \in A(i)} \exp\!\big(z_i^\top z_a / \tau_{\mathrm{con}}\big)},
    \label{eq:supcon}
\end{equation}
where $P(i)=\{p\neq i : y_p = y_i\}$ is the set of same-label
positives for anchor $i$, $A(i)=\{1,\ldots,B\}\!\setminus\!\{i\}$ is
the set of contrast samples, and $\tau_{\mathrm{con}}$ is the
contrastive temperature. We use $\lambda_1 = 0.2$ for the auxiliary
SNN BCE term and $\lambda_2 = 0.3$ for the supervised contrastive
term.

\paragraph{Surrogate gradients and auxiliary supervision.}
The standard binary LIF equations are given in
Appendix~\ref{app:lif_forward}. MAST uses the $L{=}4$ multispike forward function in
Eq.~\ref{eq:multispike_forward}, with five discrete levels,
and applies an ATan surrogate during backpropagation in the
PerChannelLIF~\cite{fang2023spikingjelly}. The surrogate does not
change the discrete forward activations. The auxiliary BCE directly supervises
the SDTB features, while the supervised contrastive term encourages
same-label clips to have similar fused representations.

\subsection{Initialization and Stability}
The per-channel LIF parameters $\tau_c$ and $V_{\mathrm{th},c}$ are
learnable for every channel $c$ of the SDTB and are parameterized in
the log domain, $\tau_c = \tau_{0}\exp(\theta_{\tau,c})$ and
$V_{\mathrm{th},c} = V_{\mathrm{th},0}\exp(\theta_{V,c})$ with
$\tau_{0} = 2.0$ and $V_{\mathrm{th},0} = 1.0$. At
initialization, $\theta_{\tau,c}$ and $\theta_{V,c}$ are zero up to
a small random perturbation, so every channel begins training near
the base values. To keep both quantities in a stable range we clamp
them at every forward pass, $\tau_c \in [0.5, 20.0]$ and
$V_{\mathrm{th},c} \in [0.05, 10.0]$. The lower bound on $\tau_c$
is a numerical safeguard only: in our trained models $\tau_c$ stays
above $1$, so the leak factor $1 - 1/\tau_c$ lies in $(0,1)$ and each
neuron acts as a decaying memory. The bounds on $V_{\mathrm{th},c}$
prevent saturate-everything and silence-everything failure modes. To keep the SDTB in a regime where
the auxiliary BCE gradient remains informative we add a small
spike-rate penalty
\begin{equation}
    \mathcal{L}_{\mathrm{rate}} \;=\; \lambda_{\mathrm{rate}}\,(\bar s - r^{*})^{2},
    \qquad r^{*} = 0.15,\ \ \lambda_{\mathrm{rate}} = 0.01,
    \label{eq:rate_penalty}
\end{equation}
where $\bar s = \mathbb{E}_{c, t}[s_t^{(c)} / L]$ is the batch-mean
normalized firing rate. The target $r^{*} = 0.15$ lies in the
$0.1$--$0.3$ range typical of recent spike-driven
transformers~\cite{yao2025sdtv3, zhou2022spikformer}. In the reported
runs this quantity is monitored during training but is not added to
the optimized objective, so the firing rate is shaped by the task
losses alone. No manual or no-grad threshold update
is performed at any point during training. Gradients are clipped by
global $L_2$ norm $1.0$, the multispike level is fixed at $L = 4$,
and the surrogate sharpness at $\alpha = 2.0$ throughout training.

\paragraph{Pseudo-event threshold parameters.}
The pseudo-event soft threshold of
Eq.~\eqref{eq:pseudo_event} is parameterized by two scalars: a
threshold level $c_{\mathrm{th}}$ and a temperature $\beta$. Both are
\emph{global} across channels and spatial locations rather than
learned per channel; channel-specific scaling is provided by the
learnable scales of Appendix~\ref{app:channels}, and temporal
adaptation by PerChannelLIF. We fix
$c_{\mathrm{th}} = 0.10$ and
$\beta = \max(c_{\mathrm{th}} \cdot 0.25,\, 10^{-6}) = 0.025$
throughout all experiments. Holding $(c_{\mathrm{th}}, \beta)$ fixed
keeps the front-end comparable across generators and matches the
role of the contrast threshold in physical event cameras
(Eq.~\eqref{eq:event_cam}), which is a fixed hardware-level
constant rather than a learned parameter.

 \subsection{Per-Channel LIF Dynamics}
The temporal integrator inside the SDTB is a per-channel
Leaky-Integrate-and-Fire (LIF) neuron with independent learnable
time constant $\tau_c$ and firing threshold $V_{\mathrm{th},c}$
for each channel $c$. Both use the log-domain parameterization of the previous
subsection, which keeps them strictly positive. The forward dynamics for channel $c$ at
timestep $t$ are
\begin{equation}
    v_t^{(c)} \;=\; \Big(1 - \tfrac{1}{\tau_c}\Big)\, v_{t-1}^{(c)} \;+\; x_t^{(c)},
    \qquad
    s_t^{(c)} \;=\; \mathrm{Spike}\!\big(v_t^{(c)}, V_{\mathrm{th},c}\big),
    \label{eq:lif_perchannel}
\end{equation}
followed by a soft reset that subtracts the emitted spike value,
$v_t^{(c)} \leftarrow v_t^{(c)} - (s_t^{(c)}/L)\, V_{\mathrm{th},c}$,
where $s_t^{(c)}/L$ is the normalized multispike output defined
below. The soft reset is used throughout.

\paragraph{Multispike forward and ATan surrogate gradient.}
Following SDT-V3~\cite{yao2025sdtv3}, the firing function is an
$L{=}4$ multispike with five levels rather than a single binary spike:
\begin{equation}
    s_t^{(c)} \;=\; \Big\lfloor\, \mathrm{clamp}\!\big(v_t^{(c)} / V_{\mathrm{th},c},\, 0,\, L\big) + 0.5 \,\Big\rfloor
    \;\in\; \{0,1,2,3,4\}, \qquad L = 4.
    \label{eq:multispike_forward}
\end{equation}
Equation~\eqref{eq:multispike_forward} is non-differentiable, so
during the backward pass we replace its derivative with an ATan
surrogate gradient~\cite{neftci2019surrogate, fang2023spikingjelly}
summed over the $L$ integer thresholds $k+\tfrac{1}{2}$ for
$k=0,\dots,L-1$:
\begin{equation}
    \frac{\partial s_t^{(c)}}{\partial u_t^{(c)}}
    \;=\;
    \mathbf{1}\!\big[0 < u_t^{(c)} < L\big]
    \sum_{k=0}^{L-1}
    \frac{\alpha/2}{1 + \big(\,\pi (u_t^{(c)} - k - \tfrac{1}{2})\,\alpha/2\,\big)^2},
    \label{eq:multispike_surrogate}
\end{equation}
where $u_t^{(c)} = v_t^{(c)}/V_{\mathrm{th},c}$ is the normalized
potential and $\alpha = 2.0$ is the sharpness. The clipping indicator
zeroes the gradient outside the saturating range, and placing one
ATan kernel at each integer threshold ensures non-zero gradient
even at the integer points where the floor function is locally
constant in Eq.~\eqref{eq:multispike_forward}.

\subsection{LIF Dynamics and Algorithm}
\label{app:lif_forward}

For reference, the standard binary Leaky
Integrate-and-Fire (LIF) neuron has discrete-time dynamics
\begin{align}
v_t &= \big(1 - \tfrac{1}{\tau}\big)\, v_{t-1} \;+\; x_t, \label{eq:lif_v}\\
s_t &= \Theta(v_t - V_{\mathrm{th}}), \label{eq:lif_s}\\
v_t &\leftarrow v_t - s_t \cdot V_{\mathrm{th}}, \label{eq:lif_reset}
\end{align}
where $v_t \in \mathbb{R}$ is the membrane potential at timestep
$t$, $x_t \in \mathbb{R}$ is the input current,
$\tau \in \mathbb{R}_{>0}$ is the membrane time constant,
$V_{\mathrm{th}} \in \mathbb{R}_{>0}$ is the firing threshold,
$\Theta(\cdot)$ is the Heaviside step function, and
$s_t \in \{0,1\}$ is the binary output spike. Eq.~\ref{eq:lif_reset}
implements a soft reset that subtracts the fired spike level
from the membrane potential after firing. Crucially, $\tau$ controls the effective temporal integration window, motivating the per-channel learnable variant used in the main text.

\begin{figure}[!t]
\centering
\begin{minipage}{0.78\linewidth}
\small
\setlength{\abovedisplayskip}{1pt}
\setlength{\belowdisplayskip}{1pt}
\hrule
\refstepcounter{algorithm}
\textbf{Algorithm \thealgorithm: MAST forward}\label{alg:mast}
\hrule
\noindent\textbf{Input} $x=(F_1,\ldots,F_T)$ and training label $y$.
\begin{enumerate}
\setlength{\itemsep}{1pt}
\setlength{\topsep}{2pt}
\item Extract clip, frame, and patch embeddings with frozen X-CLIP.
\item Construct five pixel residuals and two patch-trajectory channels
and form the pseudo-event input as described in Section~\ref{sec:method}.
\item Integrate each channel with PerChannelLIF and pass the resulting
features through the temporal module to obtain $G_t$ and the anomaly trace.
\item Project the maps and trace to $F_{\mathrm{gate}}$ and $F_{\mathrm{anom}}$.
Compute $F_{\mathrm{traj}}$ from the frame embeddings
(Appendix~\ref{app:trajectory_summary}).
\item Fuse $Z_{\mathrm{video}}$, $F_{\mathrm{anom}}$,
$F_{\mathrm{gate}}$, and $F_{\mathrm{traj}}$ and classify the clip.
During training, evaluate the objective in Eq.~\ref{eq:loss_appendix}.
\end{enumerate}
\hrule
\end{minipage}
\end{figure}

 \subsection{Implementation Notes}
\label{app:impl_notes}

The implementation of MAST used for the reported results differs
from the simplified description in the main text in the following
details. None of them changes the inputs, the model selection, or
the evaluation protocol.

\begin{enumerate}[leftmargin=1.6em,itemsep=0.25em,topsep=0.3em]
\item \textbf{Objective.} The spike-rate term of
Eq.~\ref{eq:rate_penalty} is monitored during training but is not
added to the optimized objective; the models are trained with the
three task losses of Eq.~\ref{eq:loss_appendix}.

\item \textbf{Anomaly accumulator.} The accumulator is a non-firing
LIF neuron with update
$\mathcal{A}_t = (1 - 1/\tau_{\mathrm{anom}})\,\mathcal{A}_{t-1}
+ g_t/\tau_{\mathrm{anom}}$ and $\tau_{\mathrm{anom}} = 2$, that is,
an exponential moving average. It differs from main-text
Eq.~\eqref{eq:anomaly} in its decay factor and input gain. Its input
$g_t$ is a learned non-negative combination of the spatial mean,
maximum, and standard deviation of $G_t$ rather than the mean alone.

\item \textbf{Attention scope.} Attention is computed among the
$N = 196$ spatial tokens of each frame, with time treated as a batch
dimension. Temporal integration is carried out by the PerChannelLIF
at the input and by the anomaly accumulator at the output.

\item \textbf{Gate statistics.} $F_{\mathrm{gate}}$ is computed from
three per-frame statistics of $G_t$: the spatial mean, maximum, and
standard deviation.

\item \textbf{Trajectory statistics.} The second-order sequence in
$F_{\mathrm{traj}}$ is the signed difference of successive distances
(Appendix~\ref{app:trajectory_summary}), so its statistics retain
whether the step length increases or decreases.

\item \textbf{Block composition.} Each of the eight blocks consists
of spike-driven self-attention and an MLP built from $1 \times 1$
convolutions. The convolutional stages preceding the blocks are the
per-channel $3 \times 3$ projection and the $1 \times 1$ channel
fusion (Table~\ref{tab:arch_spec}); the separable-convolution
components drawn in the main-text overview figure are not part of
the implementation.

\item \textbf{Multispike output.} Following
SDT-V3~\cite{yao2025sdtv3}, the level of
Eq.~\ref{eq:multispike_forward} is emitted in normalized form,
$s_t^{(c)}/L \in \{0, \tfrac14, \tfrac12, \tfrac34, 1\}$, and the soft
reset subtracts $(s_t^{(c)}/L)\,V_{\mathrm{th},c}$. The ATan surrogate
of Eq.~\ref{eq:multispike_surrogate} is used in the PerChannelLIF; the
multispike activations inside the blocks use a rectangular
straight-through estimator.
\end{enumerate}

\section{Additional Visualization Results}
\label{app:gatemap}
\subsection{Boundary Firing}
\begin{figure}[t]
\centering
\includegraphics[width=0.55\linewidth]{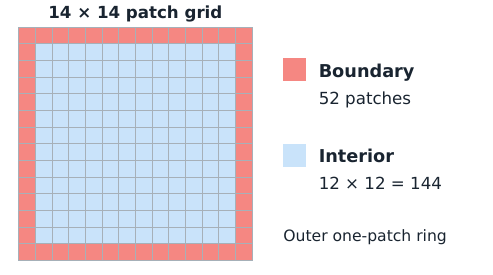}
\caption{Boundary-vs-interior partition of the $14\times14$
patch grid. The outer $1$-ring of $52$ patches forms the
\emph{boundary} mask; the inner $12\times12 = 144$ patches form
the \emph{interior} mask.}
\label{fig:boundary_def}
\end{figure}

We quantify the qualitative boundary-firing pattern of
Figures~\ref{fig:temporal_evolution_part1}--\ref{fig:temporal_evolution_part2}
by partitioning the $14\times14$ patch grid (corresponding to a
$224\times224$ input) into a \emph{boundary} ring and an
\emph{interior} block (Figure~\ref{fig:boundary_def}): the
boundary is the outer $1$-ring of $52$ patches and the interior
is the inner $12\times12 = 144$ patches. A patch is active when
its gate value exceeds the per-frame $70$th-percentile
threshold; a percentile rather than a fixed threshold is used
because per-clip gate magnitudes vary in $[0.05, 0.50]$, so a
fixed value would label entire frames as active for some clips
and entirely inactive for others. Within each frame we count
patches that participate in a run of three or more consecutive
active patches along the four boundary edges (boundary fire,
$\mathrm{BF}$) and along rows and columns of the interior block
(interior fire, $\mathrm{IF}$); both counts are normalized by
their region size ($52$ and $144$).

\paragraph{Per-generator boundary fire.}
Across the eight reported fake sources, $\mathrm{BF}$ is
$1.26$--$1.89\times$ the real-video value
(Table~\ref{tab:bf_per_generator}), whereas $\mathrm{IF}$ stays
within $0.95$--$1.35\times$ of it. The difference between real and
generated clips is therefore concentrated on the image-border ring.
These statistics describe where the gate responds; they do
not establish that the highlighted regions causally determine the
prediction.

\begin{table}[H]
\centering
\caption{Per-generator boundary fire ($\mathrm{BF}$) and
interior fire ($\mathrm{IF}$) on the GenVideo test split, at
the $70$th-percentile gate threshold; rates are fractions of
the region size ($52$, $144$).}
\label{tab:bf_per_generator}

{\small
\setlength{\tabcolsep}{6pt}
\renewcommand{\arraystretch}{1.2}
\begin{tabular}{lcc}
\toprule
\textbf{Source} & $\mathrm{BF}$ & $\mathrm{IF}$ \\
\midrule
Real         & $0.027$ & $0.141$ \\
\midrule
WildScrape   & $0.043$ & $0.134$ \\
Lavie        & $0.042$ & $0.190$ \\
HotShot      & $0.034$ & $0.165$ \\
MorphStudio  & $0.042$ & $0.142$ \\
Gen2         & $0.038$ & $0.156$ \\
ModelScope   & $0.042$ & $0.158$ \\
MoonValley   & $0.043$ & $0.168$ \\
Crafter      & $0.051$ & $0.162$ \\
\bottomrule
\end{tabular}
}
\end{table}

\subsection{Temporal Evolution}
\label{app:gatemap_temporal}

\begin{figure*}[p]
\centering
\bigfig{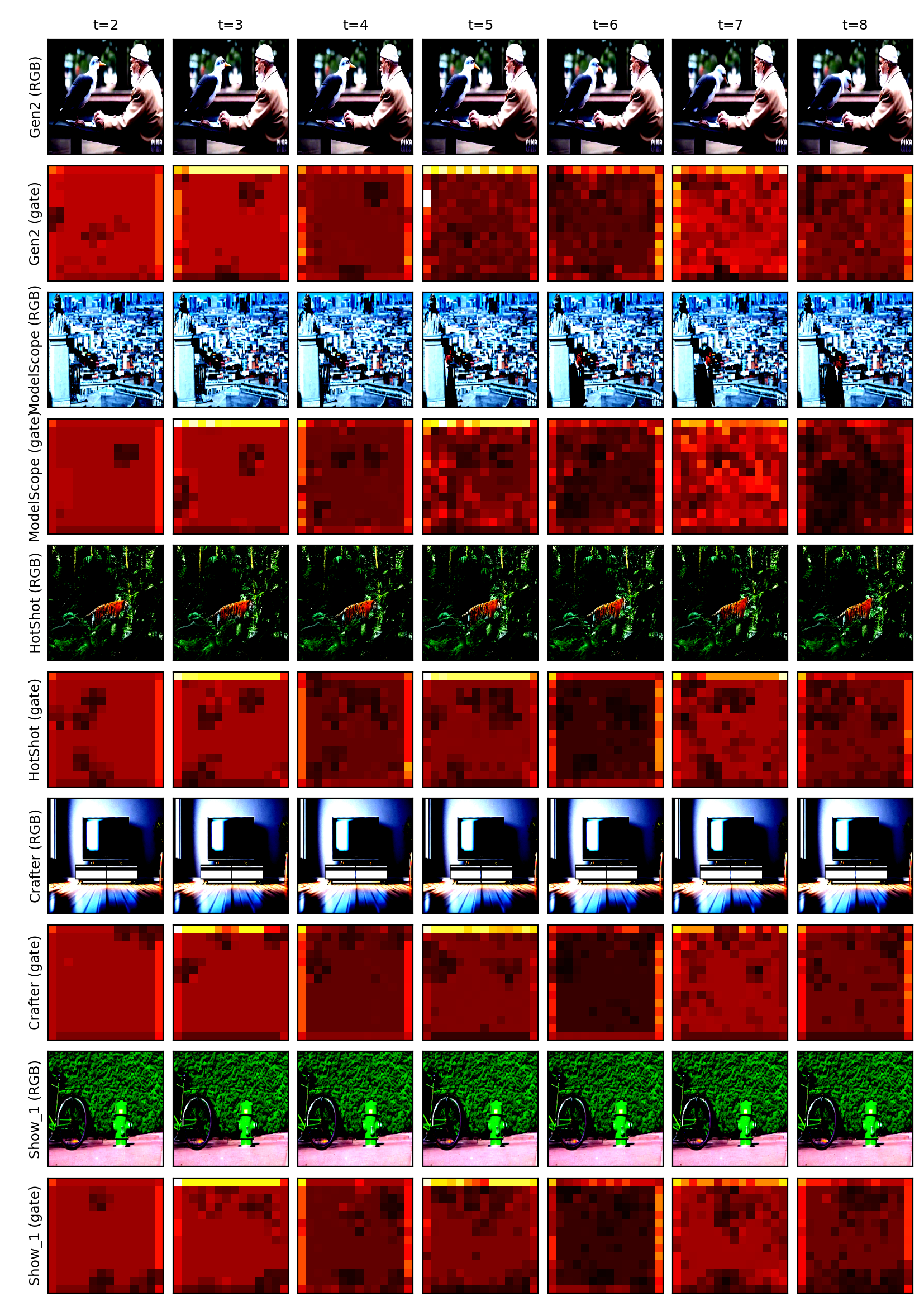}
\caption{Per-frame spike gate-map visualization on one
selected clip per GenVideo test generator (part 1 of 2:
Gen2, ModelScope, HotShot, Crafter, Show1). For each generator,
the top row shows the input RGB frames at $t=2$ to $t=8$ and
the bottom row shows the corresponding SDTB gate activation
as a heatmap. Row labels report the clip-mean gate response
$\bar A$.}
\label{fig:temporal_evolution_part1}
\end{figure*}

\begin{figure*}[p]
\centering
\bigfig{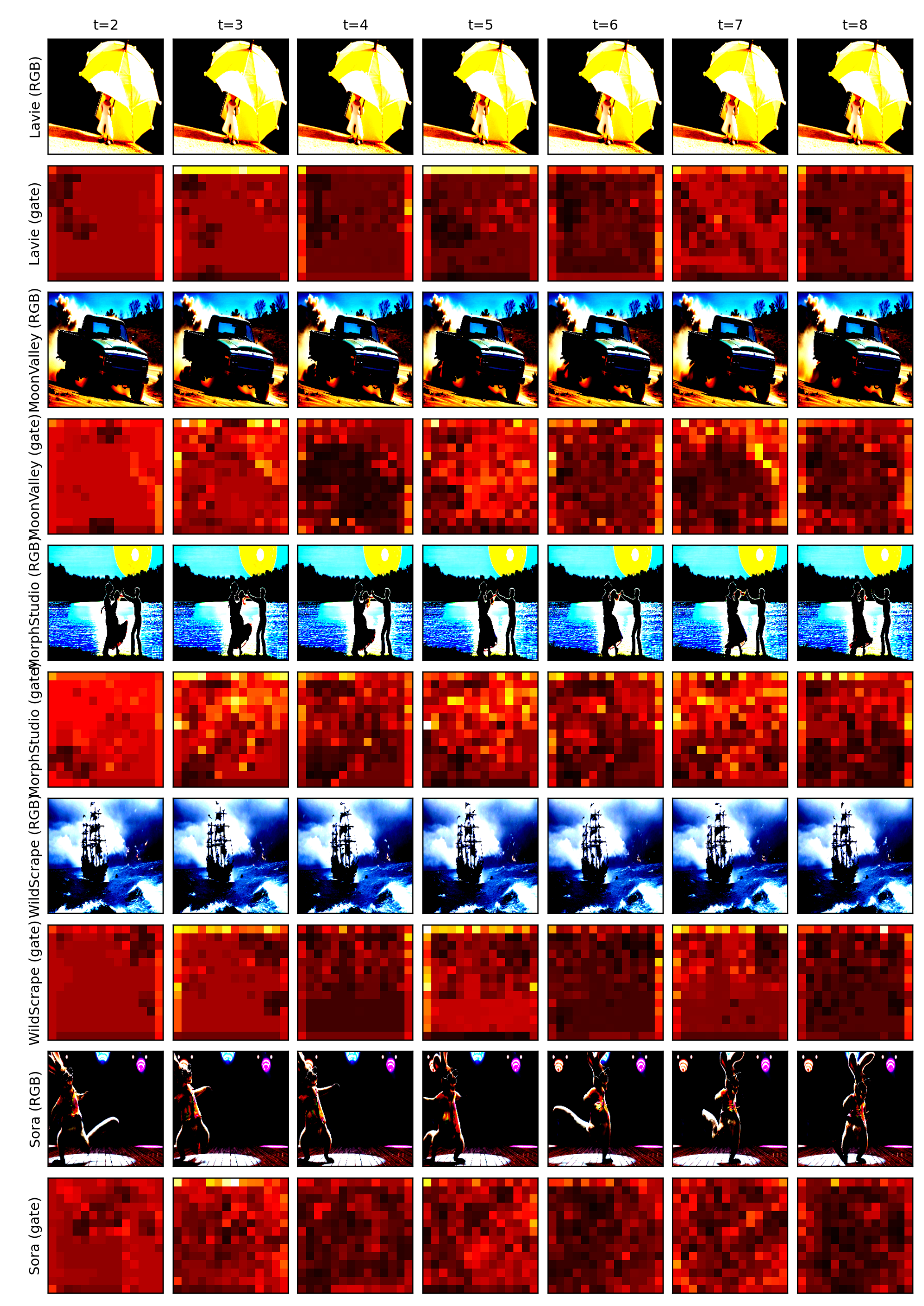}
\caption{Per-frame spike gate-map visualization on one
selected clip per GenVideo test generator (part 2 of 2:
Lavie, MoonValley, MorphStudio, WildScrape, Sora). Layout and
conventions follow Figure~\ref{fig:temporal_evolution_part1}.}
\label{fig:temporal_evolution_part2}
\end{figure*}

Figures~\ref{fig:temporal_evolution_part1} and~\ref{fig:temporal_evolution_part2}
pair RGB frames with SDTB gate maps for one selected clip from each
of ten GenVideo sources. They illustrate how the response changes across
frames within each clip. The first frame has no preceding residual.

\subsection{Residual Gallery}
\label{app:gatemap_gallery}

\begin{figure*}[p]
\centering
\bigfig[trim={0 0 0 0.7cm},clip]{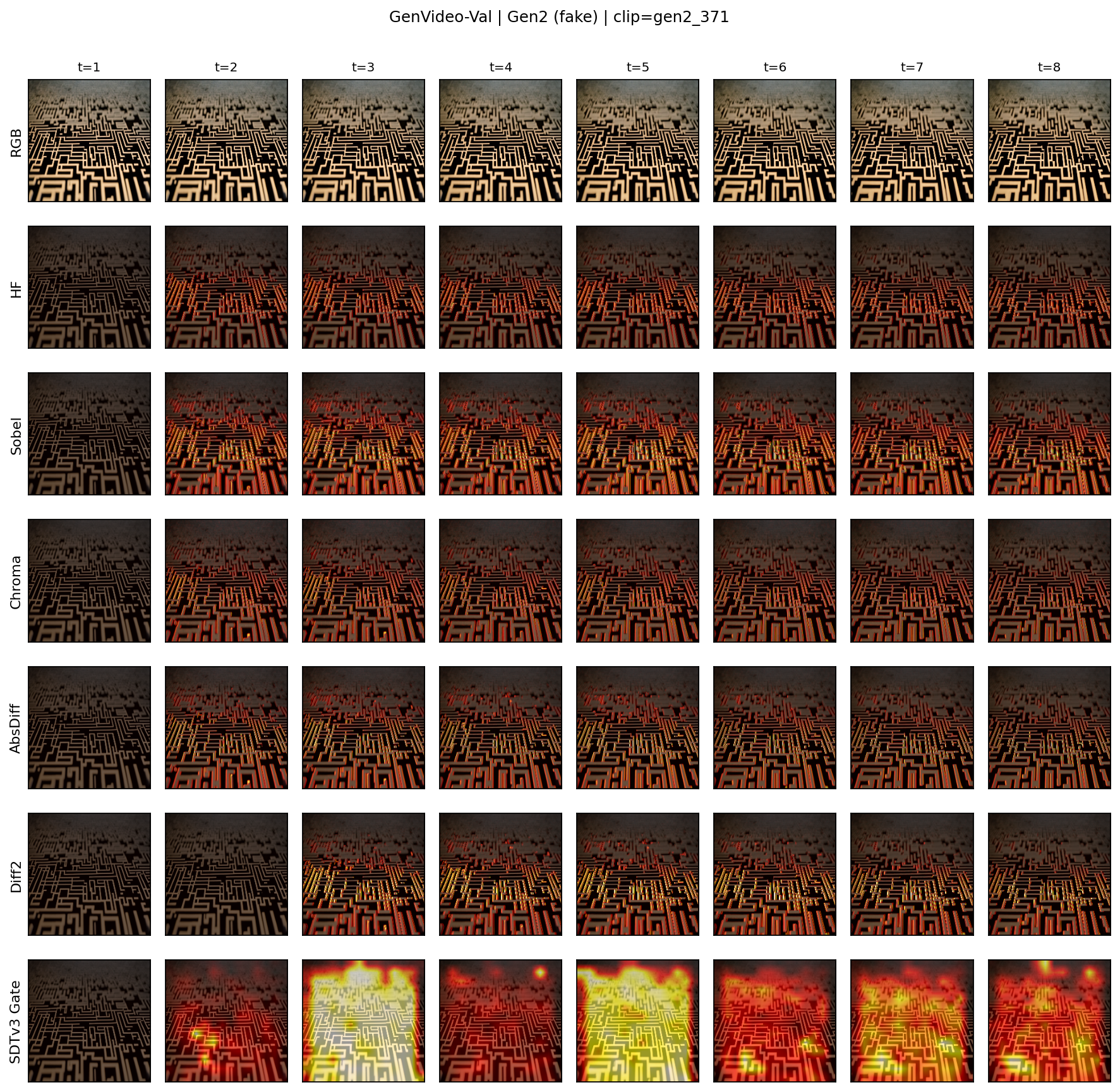}
\caption{Temporal residual channels on a fake clip generated by
Runway Gen-2, a commercial diffusion-based text-to-video model
(clip \texttt{gen2\_371}). From top to bottom: input RGB sequence,
HF, Sobel, Chroma, AbsDiff,
Diff2, and the SDTB gate map. Columns
correspond to frames $t=1$ to $t=8$.}
\label{fig:gallery_gen2}
\end{figure*}

\begin{figure*}[p]
\centering
\bigfig[trim={0 0 0 0.7cm},clip]{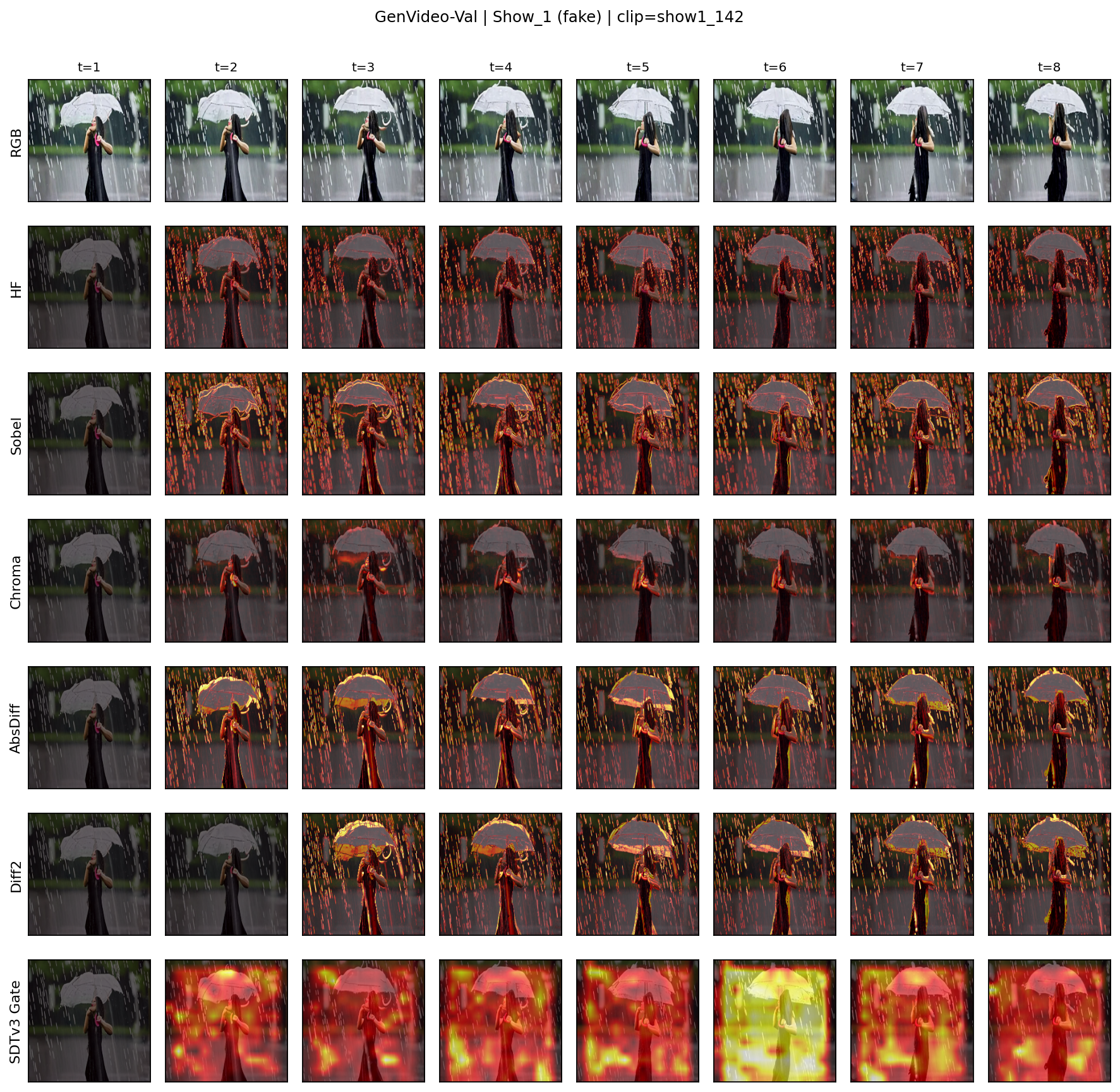}
\caption{Temporal residual channels on a fake clip generated by
Show1, a hybrid pixel-and-latent diffusion text-to-video model
(clip \texttt{show1\_142}). From top to bottom: input RGB sequence,
HF, Sobel, Chroma, AbsDiff,
Diff2, and the SDTB gate map. Columns
correspond to frames $t=1$ to $t=8$.}
\label{fig:gallery_show1}
\end{figure*}

Figures~\ref{fig:gallery_gen2} and~\ref{fig:gallery_show1} show two
selected fake clips with different motion patterns. Each panel aligns
the RGB sequence, the five pixel residuals, and the spatial gate. In the Gen2 clip, the HF and Sobel
responses follow edges in the maze as the camera moves.

In the Show1 example, moving rain produces broad AbsDiff and Diff2
responses across the frame.

\section{Hyperparameters}
\label{app:params}

We train MAST for $10$ epochs with AdamW (weight decay $0.01$) and a cosine
LR schedule, using batch size $16$ per GPU on $4$ NVIDIA RTX $3090$ GPUs
(effective batch $64$) under DDP with AMP and gradient clipping at $L_2$
norm $1.0$. The X-CLIP backbone is kept frozen throughout training.
All trainable parameters use a learning rate of $10^{-5}$.
We apply label smoothing of $0.1$. Clips are sampled as $8$ frames at
$8$ fps; frames are resized to $224 \times 224$ and scaled to $[0,1]$
without mean/std normalization. Training augmentation applies a
clip-level brightness and contrast scaling, each drawn from
$[0.8, 1.2]$, and, with probability $0.3$, additive Gaussian noise
shared across the frames of the clip. Predictions are thresholded at
$0.5$.
The objective uses $\lambda_1{=}0.2$ for the auxiliary BCE on
$\hat{y}_{\mathrm{snn}}$, $\lambda_2{=}0.3$ for the supervised
contrastive loss ($\tau_{\mathrm{con}}{=}0.07$); the spike-rate term
is monitored only (Appendix~\ref{app:impl_notes}).
Following NSG-VD~\cite{zhang2025nsgvd}, main-text
Table~\ref{tab:genvideo_pika} reports a single model, the one with the
highest validation AUROC. The module comparison in
main-text Table~\ref{tab:ablation_annsnn} reports the mean and standard
deviation over seeds 2025, 2026, and 2027.
\clearpage

\end{document}